\newif\ifarxiv
\arxivtrue

\documentclass[11pt]{article}
\usepackage{hyperref}
\hypersetup{
    colorlinks,
    linkcolor={blue!60!black},
    citecolor={blue!60!black},
    urlcolor={blue!60!black}
}
\usepackage[numbers]{natbib}
\usepackage{fullpage}

\usepackage{pgfplotstable}
\usepackage{graphicx}
\usepackage{float}
\usepackage{booktabs}

\providecommand{\comment}[1]{}

\usepackage{enumerate}
\usepackage{amssymb}
\usepackage{amsbsy}
\usepackage{amsmath}
\usepackage{amsthm}
\usepackage{amsfonts}
\usepackage{latexsym}
\usepackage{graphicx}
\usepackage{xifthen}
\usepackage{xspace}
\usepackage{bbm}
\newcommand{\cf}{{\it cf.}}
\newcommand{\eg}{{\it e.g.}}
\newcommand{\ie}{{\it i.e.}}

\newcommand{\mc}[1]{\mathcal{#1}}

\newcommand{\defeq}{:=}

\newcommand{\R}{\mathbb{R}}
\newcommand{\Z}{\mathbb{Z}}
\makeatletter
\long\def\@makecaption#1#2{
  \vskip 0.8ex
  \setbox\@tempboxa\hbox{\small {\bf #1:} #2}
  \parindent 1.5em  %
  \dimen0=\hsize
  \advance\dimen0 by -3em
  \ifdim \wd\@tempboxa >\dimen0
  \hbox to \hsize{
    \parindent 0em
    \hfil
    \parbox{\dimen0}{\def\baselinestretch{0.96}\small
      {\bf #1.} #2
    }
    \hfil}
  \else \hbox to \hsize{\hfil \box\@tempboxa \hfil}
  \fi
}
\makeatother

\newcommand{\E}{\mathbb{E}} %
\renewcommand{\P}{\mathbb{P}} %
\newcommand{\simiid}{\stackrel{\rm i.i.d.}{\sim}}

\providecommand{\argmin}{\mathop{\rm argmin}}

\providecommand{\maximize}{\mathop{\rm maximize}}

\newcommand{\floor}[1]{\left\lfloor{#1}\right\rfloor}

\newtheorem{theorem}{Theorem}

\newtheorem{proposition}[theorem]{Proposition}

\newtheorem*{claim*}{Claim}

\newenvironment{proof-sketch}{\noindent{\bf Sketch of Proof}
  \hspace*{1em}}{\qed\bigskip\\}
\newenvironment{proof-idea}{\noindent{\bf Proof Idea}
  \hspace*{1em}}{\qed\bigskip\\}

\newenvironment{proof-of-claim}{\noindent{\bf Proof of Claim}
  \hspace*{1em}}{\qed\bigskip\\}

\newenvironment{proof-of-lemma}[1][{}]{\noindent{\bf Proof of Lemma {#1}}
  \hspace*{1em}}{\qed\bigskip\\}
\newenvironment{proof-of-proposition}[1][{}]{\noindent{\bf
    Proof of Proposition {#1}}
  \hspace*{1em}}{\qed\bigskip\\}
\newenvironment{proof-of-theorem}[1][{}]{\noindent{\bf Proof of Theorem {#1}}
  \hspace*{1em}}{\qed\bigskip\\}
\newenvironment{inner-proof}{\noindent{\bf Proof}\hspace{1em}}{
  $\bigtriangledown$\medskip\\}
\newenvironment{proof-attempt}{\noindent{\bf Proof Attempt}
  \hspace*{1em}}{\qed\bigskip\\}

\newcounter{example}

\newenvironment{example*}[1][]{
  \ifthenelse{\isempty{#1}}{%
    \noindent \textbf{Example:}\hspace*{.05em}
  }{%
    \noindent \textbf{Example} ({#1})\textbf{:}\hspace*{.05em}
  }
}{%
  $\clubsuit$ \bigskip
}

\newcounter{remark}

\newenvironment{remark*}[1][]{
  \ifthenelse{\isempty{#1}}{%
    \noindent \textbf{Remark:}\hspace*{.05em}
  }{%
    \noindent \textbf{Remark} ({#1})\textbf{:}\hspace*{.05em}
  }
}{%
  $\diamondsuit$ \bigskip
}

\newcommand{\obj}{f}  %
\newcommand{\thresh}{\gamma}

\usepackage{subfigure}
\usepackage{color}
\usepackage{algorithm}
\usepackage{algpseudocode}
\usepackage{tabularx}

\usepackage{amsfonts}

\usepackage{manyfoot}

\DeclareNewFootnote{A}
\DeclareNewFootnote{B}

\let\footnoteR\footnoteB
\let\footnote\footnoteA

\usepackage{caption}
\usepackage{graphicx}

\definecolor{matlab_blue}{rgb}{0,    0.4470,   0.7410}
\definecolor{matlab_red}{rgb}{0.8500,    0.3250,    0.0980}
\definecolor{matlab_yellow}{rgb}{0.9290,    0.6940,    0.1250}
\definecolor{matlab_purple}{rgb}{0.4940,    0.1840,    0.5560}
\definecolor{matlab_green}{rgb}{0.4660,    0.6740,    0.1880}

\title{Neural Bridge Sampling for Evaluating Safety-Critical Autonomous Systems}

\makeatletter
\long\def\@makecaption#1#2{
  \vskip 0.8ex
  \setbox\@tempboxa\hbox{\small {\bf #1:} #2}
  \parindent 1.5em  %
  \dimen0=\hsize
  \advance\dimen0 by -3em
  \ifdim \wd\@tempboxa >\dimen0
  \hbox to \hsize{
    \parindent 0em
    \hfil 
    \parbox{\dimen0}{\def\baselinestretch{0.96}\small
      {\bf #1.} #2
    } 
    \hfil}
  \else \hbox to \hsize{\hfil \box\@tempboxa \hfil}
  \fi
}
\makeatother

\begin{document}
\abovedisplayskip=8pt plus0pt minus3pt
\belowdisplayskip=8pt plus0pt minus3pt

\begin{center}
  {\LARGE Neural Bridge Sampling\\for Evaluating Safety-Critical Autonomous Systems} \\
  \vspace{.5cm} {\Large Aman Sinha\footnoteR{Equal contribution}$^{1}$~~Matthew O'Kelly$^{*2}$~~Russ Tedrake$^{3}$~~John Duchi$^{1}$}\\
  \vspace{.2cm}
  $^{1}${\large Stanford University}\\
  $^2${\large University of Pennsylvania}\\
  $^3${\large Massachusetts Institute of Technology}\\
  \vspace{.2cm} \texttt{amans@stanford.edu, mokelly@seas.upenn.edu, russt@mit.edu, jduchi@stanford.edu}
\end{center}

\newif\ifarxiv
\arxivtrue
\begin{abstract}
  Learning-based methodologies increasingly find applications in
  safety-critical domains like autonomous driving and medical robotics. Due
  to the rare nature of dangerous events, real-world testing is
  prohibitively expensive and unscalable.  In this work, we employ a
  probabilistic approach to safety evaluation in simulation, where we are
  concerned with computing the probability of dangerous events. We develop a
  novel rare-event simulation method that combines exploration,
  exploitation, and optimization techniques to find failure modes and
  estimate their rate of occurrence. We provide rigorous guarantees for the
  performance of our method in terms of both statistical and computational
  efficiency. Finally, we demonstrate the efficacy of our approach on a
  variety of scenarios, illustrating its usefulness as a tool for rapid
  sensitivity analysis and model comparison that are essential to developing
  and testing safety-critical autonomous systems.
\end{abstract}

\section{Introduction}\label{sec:intro}

Data-driven and learning-based approaches have the potential to enable
robots and autonomous systems that intelligently interact with unstructured
environments. Unfortunately, evaluating the performance of
the closed-loop system is challenging, limiting
the success of such methods in safety-critical
settings. Even if we produce a deep
reinforcement learning agent better than a human at driving,
flying a plane, or performing surgery, we have no tractable way to
certify the system's quality.
Thus, currently deployed safety-critical autonomous systems are limited to
structured environments that allow mechanisms such as PID control,
simple verifiable protocols, or convex optimization to enable guarantees for
properties like stability, consensus, or recursive feasibility (see
\eg~\cite{doyle2013feedback, nakamoto2008peer,
  borrelli2017predictive}). The stylized settings of these
problems and the limited expressivity of guaranteeable properties
are barriers to solving unstructured, real-world tasks such as
autonomous navigation, locomotion, and manipulation.

The goal of this paper is to \textit{efficiently} evaluate complex systems
that lack safety guarantees and/or operate in unstructured environments. We
assume access to a simulator to test the system's performance. Given a
distribution $X\sim P_0$ of simulation parameters that describe typical
environments for the system under test, our governing problem is to estimate
the probability of an adverse event
\begin{equation}
p_\gamma \defeq \P_0(f(X) \le \gamma).
\label{eq:problem}
\end{equation}
The parameter $\gamma$ is a threshold defining an adverse event, and $f: \mc
X \to \R$ measures the safety of a realization $x$ of the agent and
environment (higher values are safer). In this work, we assume $P_0$ is
known; the system-identification and generative-modeling literatures
(\eg~\cite{aastrom1971system,
  papamakarios2019normalizing}) provide several approaches to
learn or specify
$P_0$. A major challenge for solving problem~\eqref{eq:problem}
is that the better an agent is at performing a task
(\ie~the smaller $p_{\gamma}$ is), the harder it is to confidently estimate
$p_\gamma$---one rarely observes events with $f(x) \le \gamma$.
For example, when $P_0$ is light-tailed, the sample complexity
of estimating $p_\gamma$ using naive Monte Carlo samples
grows exponentially~\cite{bucklew2013introduction}.

Problem~\eqref{eq:problem} is often solved in practice by naive Monte Carlo
estimation methods, the simplest of which \textit{explore} the search space
via random samples from $P_0$. These methods are unbiased and easy to
parallelize, but they exhibit poor sample complexity. Naive Monte Carlo can be improved
by adding an adaptive component \textit{exploiting} the most informative
portions of random samples drawn from a sequence of
approximating distributions $P_0, P_1, \dots, P_K$. However, standard
adaptive Monte Carlo methods (\eg~\cite{cerou2007adaptive}), though
they may use first-order information on the distributions $P_k$ themselves,
fail to use
first-order information about $f$ to improve sampling; we explicitly
leverage this to accelerate convergence of
the estimate through \emph{optimization}.

Naive applications of first-order optimization methods in the estimation
problem~\eqref{eq:problem}---for example biasing a sample in the direction
$-\nabla f(x)$ to decrease $f(x)$---also require second-order information to
correct for the distortion of measure that such transformations induce.  Consider the
change of variables formula for distributions $\rho(y) =
\rho(g^{-1}(y))\cdot |\det J_{g^{-1}}(y) |$ where $y=g(x)$.
When $g(x)$ is a function of the gradient $\nabla f(x)$, the volume
distortion $|\det J_{g^{-1}}(y) |$ is a function of the Hessian
$\nabla^2f(x)$. Hessian computation, if even defined, is unacceptably
expensive for high-dimensional spaces $\mc X$ and/or simulations that
involve the time-evolution of a dynamical system; our approach
avoids any Hessian computation. In contrast, gradients $\nabla f(x)$ can be efficiently computed
for many closed-loop systems~\cite{abbas2014functional,pant2017smooth,yaghoubi2018falsification, leungback} or through the use of surrogate methods~\cite{williams1992simple, deisenroth2011pilco, duchi2015optimal, BaramAnOrCasMa17}.

To that end, we propose \emph{neural bridge sampling}, a technique that
combines \textit{exploration, exploitation}, and \textit{optimization} to
efficiently solve the estimation problem~\eqref{eq:problem}. Specifically,
we consider a novel Markov-chain Monte Carlo (MCMC) scheme that moves along
an adaptive ladder of intermediate distributions $P_k$ (with corresponding
unnormalized densities $\rho_k(x)$ and normalizing constants $Z_k:=
\int_{\mc X} \rho_k(x) dx$). This MCMC scheme iteratively transforms the
base distribution $P_0$ to the distribution of interest $P_0 I\{f(x) \le
\gamma\}$. Neural bridge sampling adaptively balances exploration in the
search space (via $\nabla \log \rho_0$) against optimization (via $\nabla
f$), while avoiding Hessian computations. Our final estimate $\hat
p_{\gamma}$ is a function of the ratios $Z_k/Z_{k-1}$ of the intermediate
distributions $P_k$, the so-called ``bridges''~\cite{bennett1976efficient,
  meng1996simulating}.
We accurately estimate these ratios by warping the space between the
distributions $P_k$ using neural density estimation.

\paragraph{Contributions and outline}
Section \ref{sec:approach} presents our method, while Section
\ref{sec:analysis} provides guarantees for its statistical performance and
overall efficiency. A major focus of this work is empirical,
and accordingly, Section~\ref{sec:exp} empirically demonstrates the
superiority of neural bridge sampling over competing techniques in a variety
of applications: (i) we evaluate the sensitivity of a formally-verified
system to domain shift, (ii) we consider design optimization for
high-precision rockets, and (iii) we perform model comparisons for two learning-based approaches to autonomous navigation.

\subsection{Related Work}
\label{sec:related}
\paragraph*{Safety evaluation} Several communities \cite{corso2020survey} have attempted to evaluate the
closed-loop performance of cyber-physical, robotic, and embodied agents both
with and without learning-based components.
Existing solutions are predicated on the definition of the evaluation
problem: verification, falsification, or estimation. In this paper we
consider a method that utilizes interactions with a gradient oracle in
order to solve the estimation problem~\eqref{eq:problem}.
In contrast to our approach, the verification community has developed tools (\eg~\cite{kong2015dreach, chen2013flow, althoff2015introduction}) to investigate whether any adverse or unsafe executions of the system exist. Such methods can certify that failures are impossible, but they require that the model is written in a formal language (a barrier for realistic systems), and they require whitebox access to this formal model. Falsification approaches (\eg~\cite{esposito2004adaptive,donze2010breach,annpureddy2011s,zutshi2014multiple,dreossi2019verifai,qin2019automatic}) attempt to find \emph{any} failure cases for the system (but not the overall probability of failure). Similar to our approach, some falsification approaches (\eg~\cite{abbas2014functional,yaghoubi2018falsification}) utilize gradient information, but their goal is to simply minimize $f(x)$ rather than solve problem~\eqref{eq:problem}. Adversarial machine learning is closely related to falsification; the key difference is the domain over which the search for falsifying evidence is conducted. Adversarial examples (\eg~\cite{madry2017towards,KatzBaDiJuKo17, SinhaNaDu17,TjengTe17}) are typically restricted to a $p$-norm ball around a point from a dataset, whereas falsification considers all possible in-distribution examples. Both verification and falsification methods provide less information about the system under test than estimation-based methods: they return only whether or not the system satisfies a specification. When the system operates in an unstructured environment (\eg~driving in an urban setting), the mere existence of failures is trivial to demonstrate~\cite{shalev2017formal}. Several authors (\eg~\cite{okelly2018scalable,webb2018statistical}) have proposed that it is more important in such settings to understand the overall frequency of failures as well as the relative likelihoods of different failure modes, motivating our approach.

\paragraph*{Sampling techniques and density estimation}
When sampling rare events and estimating their probability, there are two main branches of related work: parametric adaptive importance sampling (AIS) \cite{marshall1954use,oh1992adaptive} and nonparametric sequential Monte Carlo (SMC) techniques \cite{doucet2001introduction, del2006sequential}. Both of these literatures are advanced forms of variance reduction techniques, and they are complementary to standard methods such as control variates \cite{rubinstein1985efficiency, hesterberg1998control}. Parametric AIS techniques, such as the cross-entropy method~\cite{RubinsteinKr04}, postulate a family of distributions for the optimal importance-sampling distribution. They iteratively perform heuristic optimization procedures to update the sampling distribution. SMC techniques perform sampling from a sequence of probability distributions defined nonparametrically by the samples themselves. The SMC formalism encompasses particle filters, birth-death processes, and smoothing filters \cite{del2004feynman}. Our technique blends aspects of both of these communities: we include parametric warping distributions in the form of normalizing flows \cite{papamakarios2019normalizing} within the SMC setting.

Our method employs bridge sampling \cite{bennett1976efficient, meng1996simulating}, which is closely related to other SMC techniques such as umbrella sampling \cite{chen2012monte}, multilevel splitting \cite{brehier2015analysis, cerou2007adaptive}, and path sampling \cite{gelman1998simulating}. The operational difference between these methods is in the form of the intermediate distribution used to calculate the ratio of normalizing constants. Namely, the optimal umbrella sampling distribution is more brittle than that of bridge sampling \cite{chen2012monte}. Multilevel splitting employs hard barriers through indicator functions, whereas our approach relaxes these hard barriers with smoother exponential barriers. Path sampling generalizes bridge sampling by taking discrete bridges to a continuous limit; this approach is difficult to implement in an adaptive fashion.

The accuracy of bridge sampling depends on the overlap between intermediate distributions $P_k$. Simply increasing the number of intermediate distributions is inefficient, because it requires running more simulations. Instead, we employ a technique known as \emph{warping}, where we map intermediate distributions to a common reference distribution \cite{voter1985monte,meng2002warp}. Specifically, we use normalizing flows \cite{rezende2015variational, kingma2016improved, papamakarios2017masked, papamakarios2019normalizing}, which efficiently transform arbitrary distributions to standard Gaussians through a series of deterministic, invertible functions. Normalizing flows are typically used for probabilistic modeling, variational inference, and representation learning. Recently, \citet{hoffman2019neutra} explored the benefits of using normalizing flows for reparametrizing distributions within MCMC; our warping technique encompasses this benefit and extends it to the SMC setting.

\paragraph*{Beyond simulation} This paper assumes that the generative model $P_0$ of the operating domain is given, so all failures are
in the modeled domain by definition. When deploying systems in the real world,
anomaly detection \cite{chandola2009anomaly} can discover distribution
shifts and is complementary to our approach (see \eg~\cite{choi2018waic, nachman2020anomaly}). Alternatively, the problem of distribution shift can be addressed offline via distributional robustness \cite{EsfahaniKu15,namkoong2016stochastic, rahimian2019distributionally}, where we analyze the worst-case probability of failure under an uncertainty set composed of perturbations to $P_0$.

\section{Proposed approach}\label{sec:approach}

As we note in Section \ref{sec:intro}, naive Monte Carlo measures
probabilities of rare events inefficiently. Instead, we consider a
sequential Monte Carlo (SMC) approach: we decompose the rare-event probability
$p_{\gamma}$ into a chain of intermediate quantities, each of which is
tractable to compute with standard Monte Carlo methods. Specifically,
consider $K$ distributions $P_k$ with corresponding (unnormalized)
probability densities $\rho_k$ and normalizing constants $Z_k:= \int_{\mc X}
\rho_k(x) dx$. Let $\rho_0$ correspond to the density for $P_0$ and
$\rho_\infty(x):=\rho_0(x) I\{f(x) \le \gamma\}$ be the (unnormalized)
conditional density for the region of interest. Then, we consider the
following decomposition:
\begin{equation}\label{eq:main-estimator}
	p_\gamma:=\P_0(\obj(X) \le \thresh) = \E_{P_K}\left [ \frac{Z_K}{Z_0} \frac{\rho_\infty (X)}{\rho_{K}(X)} \right ], \;\;\;\;\; \frac{Z_K}{Z_0}=\prod_{k=1}^K \frac{Z_k}{Z_{k-1}}.
\end{equation}
Although we are free to choose the intermediate distributions arbitrarily,
we will show below that our estimate for each ratio $Z_k/Z_{k-1}$ and thus
$p_\gamma$ is accurate insofar as the distributions sufficiently
overlap (a concept we make rigorous in Section~\ref{sec:analysis}). Thus,
the intermediate distributions act as bridges that iteratively steer samples
from $P_0$ towards $P_K$. One special case is the multilevel splitting
approach~\cite{kahn51,brehier2015analysis, webb2018statistical,
  norden2019efficient}, where $\rho_k(x):=\rho_0(x) I\{f(x) \le L_k\}$ for
levels $\infty=:L_0 > L_{1}\ldots >L_K:=\thresh$. In this paper, we
introduce an exponential tilting barrier~\cite{siegmund1976importance}
\begin{equation}\label{eq:rhok}
  \rho_k(x) := \rho_0(x)\exp\left ( \beta_k \left [\gamma - f(x) \right ]_-
  \right ),
\end{equation}
which allows us to take advantage of gradients $\nabla f(x)$. Here we use the ``negative ReLU'' function defined as $[x]_{-} := -[-x]_+=xI\{x<0\}$, and we assume that the measure of non-differentiable points, \eg~where $\nabla f(x)$ does not exist or $f(x)=\gamma$, is zero (see Appendix \ref{app:hmc} for a detailed discussion of this assumption). We set
$\beta_0 := 0$ and adaptively choose $\beta_k > \beta_{k-1}$. The parameter
$\beta_k$ tilts the distribution towards the distribution of interest:
$\rho_k \to \rho_{\infty}$ as $\beta_k\to \infty$. In what follows, we
describe an MCMC method that combines exploration, exploitation, and
optimization to draw samples $X_i^k \sim P_k$. We then show how to compute
the ratios $Z_k/Z_{k-1}$ given samples from both $P_{k-1}$ and
$P_{k}$. Finally, we describe an adaptive way to choose the intermediate
distributions $P_k$. Algorithm \ref{alg:main} summarizes the overall
approach.

\begin{algorithm}[t]
  \caption{\label{alg:main} Neural bridge sampling}
  \begin{small}
  \begin{algorithmic}[]
    \State \textbf{Input:} $N$ samples $x_i^0 \simiid P_0$, MCMC steps $T$, step size $\alpha \in (0,1)$, stop condition $s \in (0,1)$
    \State Initialize $k\gets 0$, $\beta_0 \gets 0$, $\log(\hat p_\gamma) \gets 0$
    \State \textbf{while} {$\frac{1}{N} \sum_{i} I\{f(x_i^k) \le \gamma \} < s$ } \textbf{do}
    \State~~~~~$\beta_{k+1} \gets$ solve problem~\eqref{eq:beta_anneal}
    \State~~~~~\textbf{for} $i=1$ to $N$, in parallel
    \State~~~~~~~~~~$x_i^{k+1} \simiid \text{Mult}(\{\rho_{k+1}(x_i^k)/\rho_{k}(x_i^k)\})$\; // multinomial resampling
    \State~~~~~\textbf{for} $t=1$ to $T$
    \State~~~~~~~~~~\textbf{for} $i=1$ to $N$, in parallel
    \State~~~~~~~~~~~~~~~$x_i^{k+1}\gets$ $\text{WarpedHMC}(x_i^{k}, \theta_k)$\;  // Appendix~\ref{app:hmc}
        \State~~~~~$\theta_{k+1} \gets \argmin \text{problem}$~\eqref{eq:flowprob}\; // train normalizing flow on $\{x_i^{k+1}\}$ via SGD
    \State~~~~~$\log(\hat p_\gamma) \gets \log(\hat p_\gamma)  + \log(Z_{k+1}/Z_k)$\; // warped bridge estimate~\eqref{eq:bridgewarp}

    \State~~~~~$k\gets k+1$
    \State $\log(\hat p_\gamma) \gets \log(\hat p_\gamma) + \log(\frac{1}{N} \sum_{i} I\{f(x_i^k) \le \gamma \})$

  \end{algorithmic}\end{small}
\end{algorithm}

\paragraph{MCMC with an exponential barrier}
Gradient-based MCMC techniques such as the Metropolis-adjusted Langevin
algorithm (MALA) \cite{rossky1978brownian,roberts2002langevin} or
Hamiltonian Monte Carlo (HMC)~\cite{duane1987hybrid,neal2012mcmc} use
gradients $\nabla \log \rho_0(x)$ to efficiently explore the space $\mc X$
and avoid inefficient random-walk behavior~\cite{durmus2017nonasymptotic,
  chen2019fast}. Classical mechanics inspires the HMC approach: HMC introduces an auxiliary random momentum variable $v \in \mc V$ and generates
proposals by performing Hamiltonian dynamics in the augmented state-space
$\mc X \times \mc V$. These dynamics conserve volume in the augmented
state-space, even when performed with discrete time steps
\cite{leimkuhler2004simulating}.

By including the barrier $\exp\left ( \beta_k \left [\gamma - f(x) \right
]_- \right )$, we combine exploration with optimization; the magnitude of
$\beta_k$ in the barrier modulates the importance of $\nabla f$
(optimization) over $\nabla \log \rho_0$ (exploration), two elements of the HMC proposal (see Appendix \ref{app:hmc} for details). We discuss the
adaptive choice for $\beta_k$ below. Most importantly, we avoid any need for
Hessian computation because the dynamics conserve volume. 
As Algorithm~\ref{alg:main} shows, we perform MCMC as follows: given $N$
samples $x_i^{k-1} \sim P_{k-1}$ and a threshold $\beta_k$, we first
resample using their importance weights (exploiting the performance of
samples that have lower function value than others) and then perform $T$ HMC
steps. In this paper, we implement split HMC \cite{shahbaba2014split} which
is convenient for dealing with the decomposition of $\log \rho_k(x)$ into
$\log \rho_0(x) + \beta_k[\gamma -f(x)]_-$~(see Appendix~\ref{app:hmc} for
details).

\paragraph{Estimating $Z_k/Z_{k-1}$ via bridge sampling}
Bridge sampling~\cite{bennett1976efficient, meng1996simulating}
allows estimating the ratio of normalizing constants of two
distributions by rewriting
\begin{equation}\label{eq:bridge}
  \small
  E_k := \frac{Z_k}{Z_{k-1}}= \frac{Z_k^B/Z_{k-1}}{Z_k^B/Z_k} = \frac{\E_{P_{k-1}} [\rho^B_k (X)/\rho_{k-1}(X) ]}{\E_{P_{k}} [\rho^B_k (X)/\rho_{k}(X)]},
  \qquad
  \widehat E_k = \frac{\sum_{i=1}^N \rho^B_k (x_i^{k-1})/\rho_{k-1}(x_i^{k-1}) }{\sum_{i=1}^N \rho^B_k (x_i^{k})/\rho_{k}(x_i^{k}) },
\end{equation}
where $\rho_k^B$ is the density for a bridge distribution between $P_{k-1}$
and $P_{k}$, and $Z_k^B$ is its associated normalizing constant. We employ
the geometric bridge $\rho_k^B(x):=\sqrt{\rho_{k-1}(x) \rho_k(x)}$. In
addition to being simple to compute, bridge sampling with a geometric bridge
enjoys the asymptotic performance guarantee that the relative mean-square error
scales inversely with the Bhattacharyya coefficient, $G(P_{k-1},
P_k)=\int_\mc{X} \sqrt{\frac{\rho_{k-1}(x)}{Z_{k-1}}\frac{\rho_k(x)} {Z_k}}dx \in [0,1]$ (see
Appendix~\ref{app:performance} for a proof). This value is closely related
to the Hellinger distance, $H(P_{k-1}, P_k)=\sqrt{2-2G(P_{k-1}, P_k)}$. In
Section \ref{sec:analysis}, we analyze the ramifications of this fact on the
overall convergence of our method.

\paragraph{Neural warping}
Both HMC and bridge sampling benefit from warping samples $x_i$ into a
different space. As \citet{betancourt2017conceptual} notes, HMC mixes
poorly in spaces with ill-conditioned
geometries. ~\citet{girolami2011riemann} and~\citet{hoffman2019neutra}
explore techniques to improve mixing efficiency by minimizing shear in the
corresponding Hamiltonian dynamics. One way to do so is to transform to a
space that resembles a standard isotropic Gaussian \cite{mangoubi2017rapid}.

Conveniently, transforming $P_k$ to a common distribution (\eg~a standard Gaussian) also benefits the bridge-sampling estimator~\eqref{eq:bridge}. As noted above, the error of the bridge estimator grows with the Hellinger distance between the distributions $H(P_{k-1}, P_k)$. However, normalizing constants $Z_k$ are invariant to (invertible) transformations. Thus, transformations that warp the space between distributions reduce the error of the bridge-sampling estimator \eqref{eq:bridge}. Concretely, we consider invertible transformations $W_{k}$ such that $y^k_i = W_{k}(x_i^k)$. For clarity of notation, we write probability densities over the space $\mc Y$ as $\phi$, the corresponding distributions for $Y^k$ as $Q_k$, and the inverse transformations $W_{k}^{-1}(y)$ as $V_k(y)$. Then we can write the bridge-sampling estimate~\eqref{eq:bridge} in terms of the transformed variables $y$. The numerator and denominator are as follows:
\begin{small}
\begin{subequations}\label{eq:bridgewarp}
\begin{equation}
\E_{Q_{k-1}} \left [\frac{\phi^B_k (Y)}{\phi_{k-1}(Y)} \right ] =  \E_{Q_{k-1}} \left [\sqrt \frac{\phi_k (Y)}{\phi_{k-1}(Y)} \right ] = 
\E_{Q_{k-1}} \left [\sqrt \frac{\rho_k (V_k(Y))|\det J_{V_k}(Y) | }{\rho_{k-1}(V_{k-1}(Y)) |\det J_{V_{k-1}}(Y) |} \right ],
\end{equation}
\begin{equation}
\E_{Q_{k}} \left [\frac{\phi^B_k (Y)}{\phi_{k}(Y)} \right ] =  \E_{Q_{k}} \left [\sqrt \frac{\phi_{k-1} (Y)}{\phi_{k}(Y)} \right ] = 
\E_{Q_{k}} \left [\sqrt \frac{\rho_{k-1} (V_{k-1}(Y))|\det J_{V_{k-1}}(Y) | }{\rho_{k}(V_{k}(Y)) |\det J_{V_{k}}(Y) |} \right ].
\end{equation}\end{subequations}\end{small}\noindent By transforming all $P_k$ into $Q_k$ to resemble standard Gaussians, we reduce the Hellinger distance $H(Q_{k-1}, Q_k) \le H(P_{k-1}, P_k)$. Note that the volume distortions in the expression~\eqref{eq:bridgewarp} are functions of the transformation $V_k$, so they do not require computation of the Hessian $\nabla^2f$. However, computing $\rho_k(V_k(y))$ requires evaluations of $f$ (\eg~calls of the simulator). We consider the cost-benefit analysis of warping in Section \ref{sec:analysis}.

Classical warping techniques include simple mean shifts or affine scaling~\cite{voter1985monte,meng2002warp}. Similar to \citet{hoffman2019neutra}, we consider normalizing flows, a much more expressive class of transformations that have efficient Jacobian computations \cite{papamakarios2019normalizing}. Specifically, given samples $x_i^k$, we train masked autoregressive flows (MAFs)~\cite{papamakarios2017masked} to minimize the empirical KL divergence between the transformed samples $y_i^k$ and a standard Gaussian $D_{\mathrm{KL}}(Q_k \| \mathcal{N}(0, I))$. Parametrizing $W_k$ by $\theta_k$, this minimization problem is equivalent to:
\begin{equation}\label{eq:flowprob}
\text{minimize}_{\theta}  \sum_{i=1}^N -\log \left |\det J_{W_{k}}\left (x_i^k; \theta \right ) \right | + \frac{1}{2} \left \|W_k \left (x_i^k; \theta \right ) \right \|_2^2.
\end{equation}
The KL divergence is an upper bound to the Hellinger distance; we found minimizing the former to be more stable than minimizing the latter. Furthermore, to improve training efficiency, we exploit the iterated nature of the problem and warm-start the weights $\theta_k$ with the trained values $\theta_{k-1}$ when solving problem~\eqref{eq:flowprob} via stochastic gradient descent (SGD). As a side benefit, the trained flows can be repurposed as importance-samplers for the ladder of distributions from nominal behavior to failure.

\paragraph{Adaptive intermediate distributions}
Because we assume no prior knowledge of the system under test, we exploit previous progress to choose the intermediate $\beta_k$ online; this is a key difference to our approach compared to other forms of sequential Monte Carlo (\eg~\cite{neal2001annealed, neal2005estimating}) which require a predetermined schedule for $\beta_k$. We define the quantities
\begin{equation}
	a_k:=\textstyle\sum_{i}^N I\{ f(x_i^k) \le \gamma \}/N, \;\; b_k(\beta) := \textstyle\sum_{i=1}^N 
	\exp \left ( (\beta - \beta_k)[\gamma - f(x_i^k)]_{-} \right )/N.
\end{equation}
The first is the fraction of samples that have achieved the threshold. The second is an importance-sampling estimate of $E_{k+1}$ given samples $x_i^k \sim P_k$, written as a function of $\beta$. For fixed fractions $\alpha, s \in (0,1)$ with $\alpha < s$, $\beta_{k+1}$ solves the following optimization problem:
\begin{equation}\label{eq:beta_anneal}
  \maximize~\beta\;\;\text{s.t.}\;\;\{ b_k(\beta) \ge \alpha, \;\;  a_k /b_k(\beta) \le s\}.
\end{equation}
Since $b_k(\beta)$ is monotonically decreasing and $b_k(\beta) \ge a_k$, this problem can be solved efficiently via binary search. The constant $\alpha$ tunes how quickly we enter the tails of $P_0$ (smaller $\alpha$ means fewer iterations), whereas $s$ is a stop condition for the last iteration. Choosing $\beta_{k+1}$ via~\eqref{eq:beta_anneal} yields a crude estimate for the ratio $Z_{k+1}/Z_{k}$ as $\alpha$ (or $a_{K-1}/s$ for the last iteration). The bridge-sampling estimate $\widehat E_{k+1}$ corrects this crude estimate once we have samples from the next distribution $P_{k+1}$.

\section{Performance analysis} \label{sec:analysis}
We can write the empirical estimator of the function~\eqref{eq:main-estimator} as
\begin{equation}\label{eq:estimator}
\hat p_\gamma = \prod_{k=1}^K \widehat E_k \frac{1}{N}\sum_{i=1}^N \frac{\rho_\infty(x^K_i)}{\rho_K(x^K_i)},
\end{equation}
where $\widehat E_k$ is given by the expression~\eqref{eq:bridge} without warping, or similarly, as a Monte Carlo estimate of the expression~\eqref{eq:bridgewarp} with warping. We provide guarantees for both the time complexity of running Algorithm \ref{alg:main} (\ie~the iterations $K$) as well as the overall mean-square error of $\hat p_\gamma$. For simplicity, we provide results for the asymptotic (large $N$) and well-mixed MCMC (large $T$) limits. Assuming these conditions, we have the following: 

\begin{proposition}\label{prop:main}
Let $K_0:=\floor{ \log(p_\gamma) / \log(\alpha) }$. Then, for large $N$ and $T$, $s\ge 1/3$, and $p_\gamma < s$, the total number of iterations in Algorithm \ref{alg:main} approaches $K \overset{\mathrm{a.s.}}{\to} K_0 + I\{p_{\gamma}/\alpha^{K_0} < s \}$. Furthermore, for the non-warped estimator, the asymptotic relative mean-square error~~$\E[\left ({\hat p_{\gamma}}/{p_{\gamma}}-1\right )^2]$~~is
\begin{small}
 \begin{equation}\label{eq:varbound}
\frac{2}{N}\sum_{k=1}^K \left (\frac{1}{G(P_{k-1}, P_k)^2}-1 \right ) -\frac{2}{N} \sum_{k=1}^{K-1} \left ( \frac{G(P_{k-1}, P_{k+1})}{G(P_{k-1}, P_{k})G(P_{k}, P_{k+1})} -1\right ) +\frac{1-s}{sN}+ o\left(\frac{1}{N}\right).
 \end{equation}
 \end{small}
In particular, if the inverse Bhattacharyya coefficients are bounded such that $\frac{1}{G(P_{k-1}, P_k)^2} \le D$ (with $D \ge 1$), then the asymptotic relative mean-square error satisfies $\E[\left ({\hat p_{\gamma}}/{p_{\gamma}}-1\right )^2] \le 2KD/N$. For the warped estimator, replace $G(P_i, P_j)$ with $G(Q_i, Q_j)$ in the expression~\eqref{eq:varbound}. 
\end{proposition}

See Appendix \ref{app:performance} for the proof. We provide some remarks about the above result. Intuitively, the first term in the bound~\eqref{eq:varbound} accounts for the variance of $\widehat E_k$. The denominator of $\widehat E_{k-1}$ and numerator of $\widehat E_k$ both depend on $x_i^k$; the second sum in~\eqref{eq:varbound} accounts for the covariance between those terms. Furthermore, the quantities in the bound~\eqref{eq:varbound} are all empirically estimable, so we can compute the mean-square error from a single pass of Algorithm \ref{alg:main}. In particular,
\begin{equation}
	G(P_{k-1}, P_k)^2 = \frac{Z^B_k}{Z_{k-1}}\frac{Z^B_k}{Z_k}, \;\;\;\;\;\;\;\;\;\;\;\;\;\;\; \frac{G(P_{k-1}, P_{k+1})}{G(P_{k-1}, P_{k})G(P_{k}, P_{k+1})} = \frac{Z^C_k}{Z_k} \frac{Z_k}{Z^B_k}\frac{Z_k}{Z^B_{k+1}},
\end{equation}
where $Z^C_k/Z_k = \E_{P_k}\left [ \rho^B_{k}(X)\rho^B_{k+1}(X)/\rho_k(X)^2 \right ]$. The last term in the bound~\eqref{eq:varbound} is the relative variance of the final Monte Carlo estimate $\sum_{i} I\{f(x_i^K) \le \gamma \}/N$. 

\paragraph{Overall efficiency}
The statistical efficiency outlined in Proposition \ref{prop:main} is pointless if it is accompanied by an overwhelming computational cost. We take the atomic unit of computation to be a query of the simulator, which returns both evaluations of $f(x)$ and $\nabla f(x)$; we assume other computations to be negligible compared to simulation. %
As such, the cost of Algorithm \ref{alg:main} is $N(1+ KT)$ evaluations of the simulator without warping and $N(1+KT) + 2KN$ with warping. Thus, the relative burden of warping is minimal, because training the normalizing flows to minimize $D_{\mathrm{KL}}(Q_k \| \mathcal{N}(0, I))$ requires no extra simulations. In contrast, directly minimizing $D_{\mathrm{KL}}(Q_{k-1} \| Q_{k})$ would require extra simulations at each training step to evaluate $\rho_k(V_k(y))$. 

Our method can exploit two further sources of efficiency. First, we can employ surrogate models for gradient computation and/or function evaluation during the $T$ MCMC steps. For example, using a surrogate model for a fraction $d\le 1-1/T$ of the MCMC iterations reduces the factor $T$ to $T_s:=(1-d)T$ in the overall cost. %
Surrogate models have an added benefit of making our approach amenable for simulators that do not provide gradients. The second source of efficiency is parallel computation. Given $C$ processors, the factor $N$ in the cost drops to $N_c:=\left \lceil{N/C}\right \rceil$.%

The overall efficiency of the estimator~\eqref{eq:estimator}---relative error multiplied by cost~\cite{hammersley1964monte}---depends on $p_\gamma$ as $\log(p_{\gamma})^2$. In contrast, the standard Monte Carlo estimator has cost $N$ to produce an estimate with relative error $\frac{1-p_{\gamma}}{p_{\gamma}N}$. Thus, the relative efficiency gain for our estimator~\eqref{eq:estimator} over naive Monte Carlo is $O(1/({p_{\gamma}\log(p_{\gamma})^2}))$: the efficiency gains over naive Monte Carlo increase as $p_{\gamma}$ decreases. %
\section{Experiments}
\label{sec:exp}

We evaluate our approach in a variety of scenarios, showcasing its use in efficiently evaluating the safety of autonomous systems. We begin with a synthetic problem to illustrate the methodology concretely as well as highlight the pitfalls of using gradients naively. Then, we evaluate a formally-verified neural network controller~\cite{ivanov2019verisig} on the OpenAI Gym continuous MountainCar environment~\cite{moore1990efficient, brockman2016openai} under a domain perturbation. Finally, we consider two examples of using neural bridge sampling as a tool for engineering design in high-dimensional settings: (a) comparing thruster sizes to safely land a rocket~\cite{blackmore2017autonomous} in the presence of wind, and (b) comparing two algorithms on the OpenAI Gym CarRacing environment (which requires a surrogate model for gradients)~\cite{klimovcarracing}.

We compare our method with naive Monte Carlo (MC) and perform ablation studies for the effects of neural warping (denoted as NB with warping and B without). We also provide comparisons with adaptive multilevel splitting (AMS) \cite{brehier2015analysis, webb2018statistical, norden2019efficient}. All methods are given the same computational budget as measured by evaluations of the simulator. This varies from 50,000-100,000 queries to run Algorithm \ref{alg:main} as determined by $p_{\gamma}$ (see Appendix \ref{app:appendix_experiments} for details of each experiment's hyperparameters). However, despite running Algorithm \ref{alg:main} with a given $\gamma$, we evaluate estimates $\hat p_{\gamma_{\mathrm{test}}}$ for all $\gamma_{\mathrm{test}}\ge\gamma$. Larger $\gamma_{\mathrm{test}}$ require fewer queries to evaluate $\hat p_{\gamma_{\mathrm{test}}}$ (as Algorithm \ref{alg:main} terminates early). Thus, we adjust the number of MC queries accordingly for each $\gamma_{\mathrm{test}}$. Independently, we calculate the ground-truth values $p_{\gamma_{\mathrm{test}}}$ for the non-synthetic problems using a fixed, very large number of MC queries. 

\paragraph{Synthetic problem} We consider the two-dimensional function $f(x)=-\min(|x_{[1]}|, x_{[2]})$, where $x_{[i]}$ is the $i$\textsuperscript{th} dimension of $x \in \R^2$. We let $\gamma\!=\!-3$ and $P_0\!=\!\mathcal{N}(0,I)$ (for which $p_{\gamma}=3.6\cdot 10^{-6}$). Note that $\nabla^2 f(x)=0$ almost everywhere, yet $\nabla f(x)$ has negative divergence in the neighborhoods of $x_{[2]}\!=\!|x_{[1]}|$. Indeed, gradient descent collapses $x_i\!\sim\!P_0$ to the lines $x_{[2]}\!=\!|x_{[1]}|$, and the ill-defined nature of the Hessian makes it unsuitable to track volume distortions. Thus, simple gradient-based transformations used to find adversarial examples (\eg~minimize $f(x)$) should not be used for estimation in the presence of non-smooth functions, unless volume distortions can be quantified.

\begin{figure}[!!t]
\begin{minipage}{0.32\columnwidth}
\centering
\subfigure[Samples colored by iteration]{\label{fig:toy-a}\includegraphics[width=0.9\textwidth]{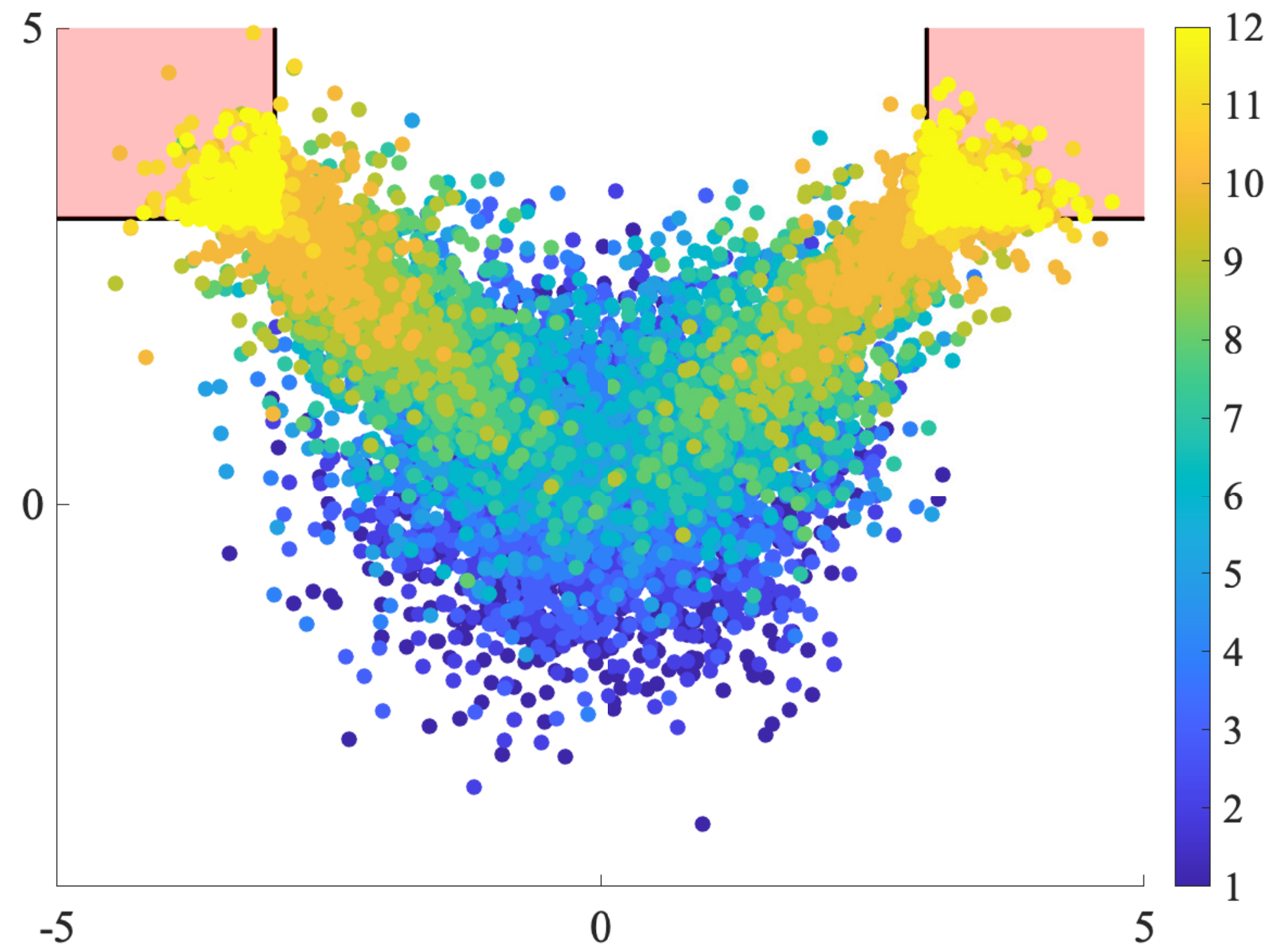}}%
\end{minipage}
\begin{minipage}{0.32\columnwidth}
\centering
\subfigure[$\hat p_{\gamma_{\mathrm{test}}}$ vs. $\gamma_{\mathrm{test}}$]{\label{fig:toy-b}\includegraphics[width=0.9\textwidth]{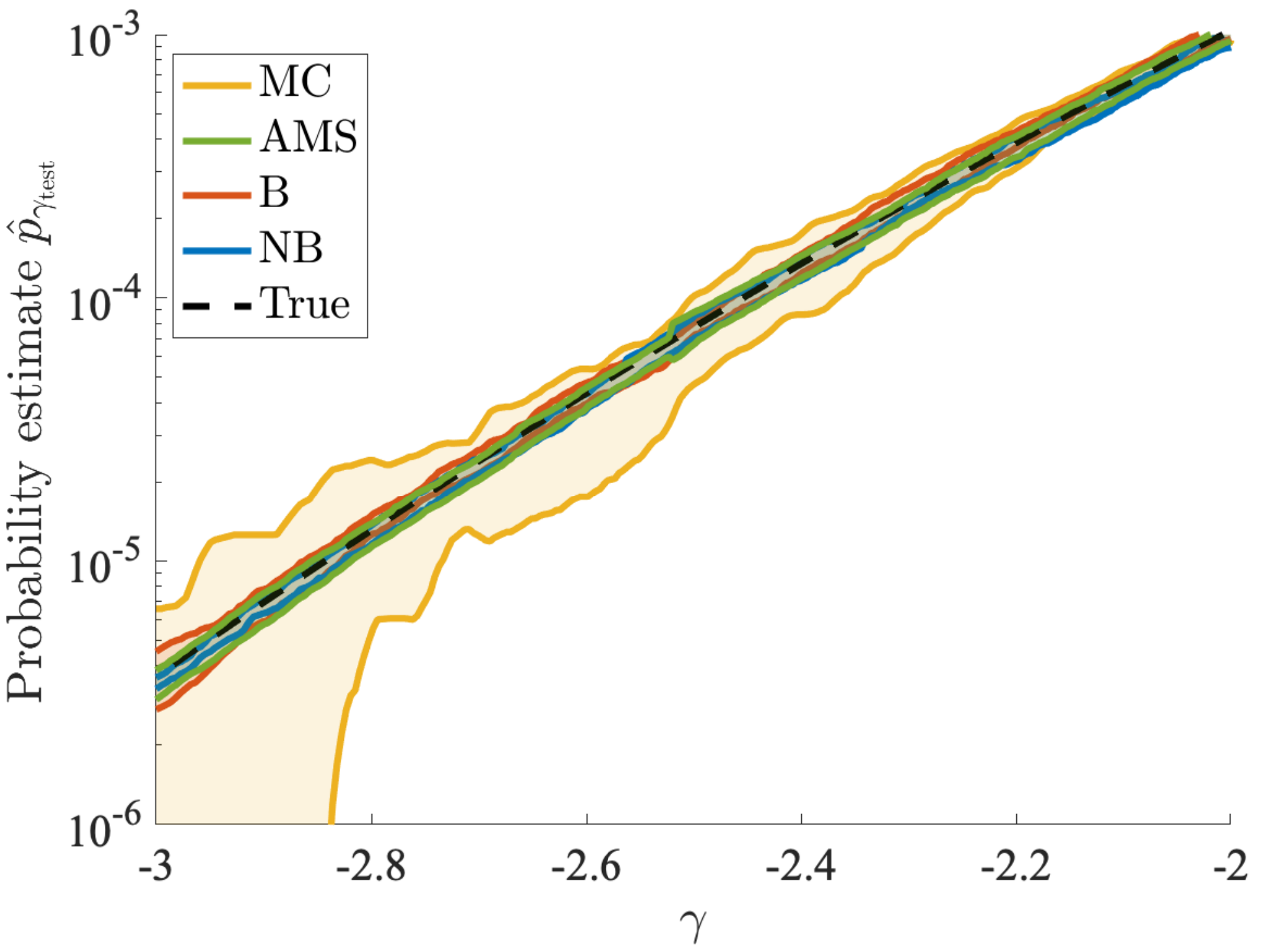}}%
\end{minipage}
\begin{minipage}{0.32\columnwidth}
\centering
\subfigure[Ratio of variance vs. $p_{\gamma_{\mathrm{test}}}$]{\label{fig:toy-c}\includegraphics[width=0.9\textwidth]{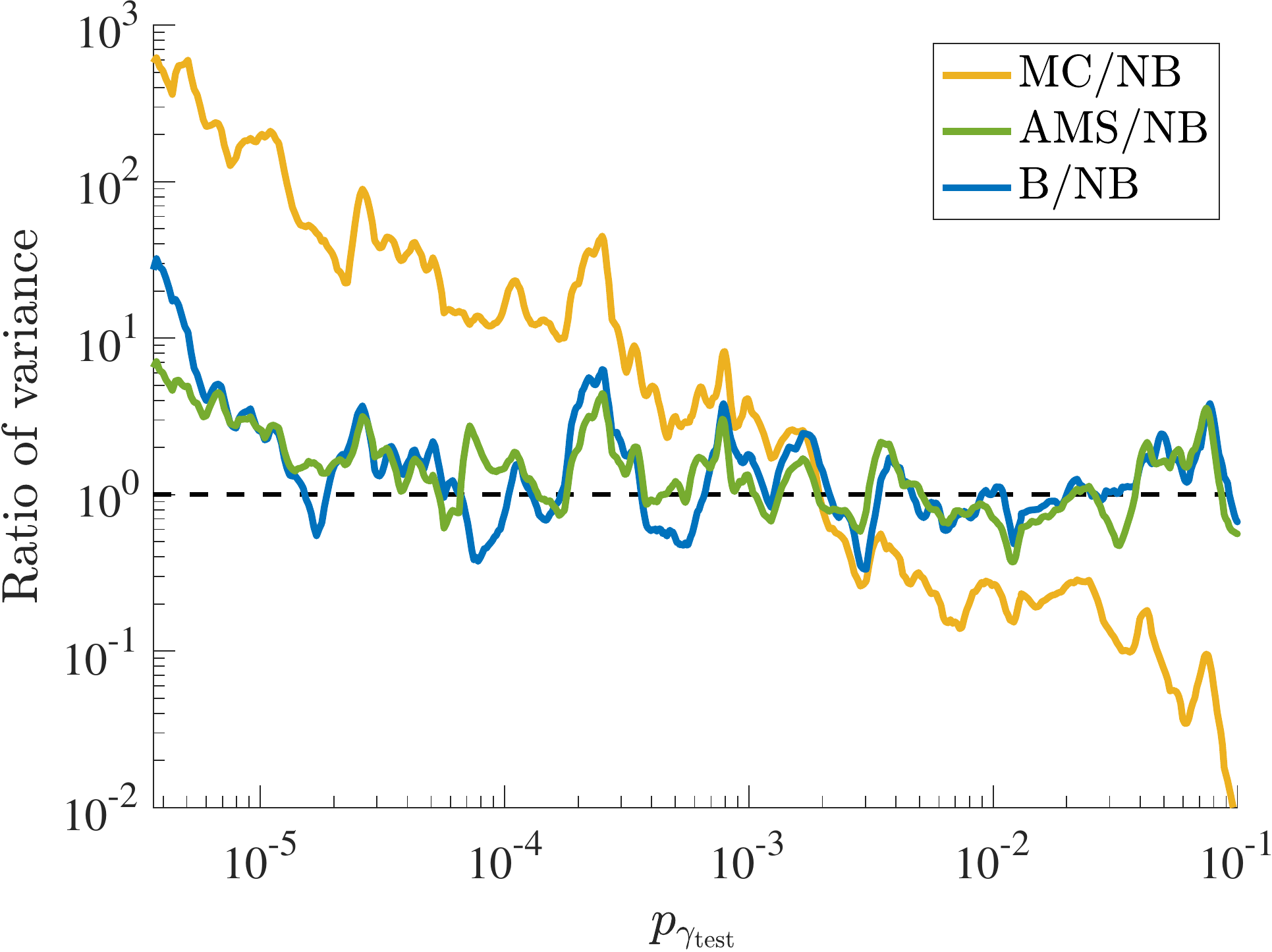}}%
\end{minipage}
\centering
\caption[]{\label{fig:toy}Experiments on a synthetic problem. 10 trials are used to calculate the 99\% confidence intervals in (b) and variance ratios in (c). All adaptive methods perform similarly in this well-conditioned search space except at very small $\gamma$, where NB performs the best. 
	}
	\vskip -10pt
\end{figure}

Figure \ref{fig:toy-a} shows the region of interest in pink and illustrates the gradual warping of $\rho_0$ towards $\rho_{\infty}$ over iterations of Algorithm \ref{alg:main}. Figures \ref{fig:toy-b} and \ref{fig:toy-c} indicate that all adaptive methods outperform MC for $p_{\gamma_{\mathrm{test}}}<10^{-3}$. For larger $p_{\gamma_{\mathrm{test}}}$, the overhead of the adaptive methods renders MC more efficient (Figure \ref{fig:toy-c}). The linear trend of the yellow MC/NB line in Figure \ref{fig:toy-c} aligns with the theoretical efficiency gain discussed in Section \ref{sec:analysis}. Finally, due to the simplicity of the search space and the landscape of $f(x)$, the benefits of gradients and warping are not drastic. Specifically, as shown in Figure \ref{fig:toy-c}, all adaptive methods have similar confidence in their estimates except at very small $p_{\gamma_{\mathrm{test}}}<10^{-5}$, where NB outperforms AMS and B. The next example showcases the benefits of gradients as well as neural warping in a more complicated search space.

\begin{figure}[!!t]
\begin{minipage}{0.32\columnwidth}
\centering
\subfigure[The environment]{\label{fig:mountain-a}\includegraphics[width=0.9\textwidth]{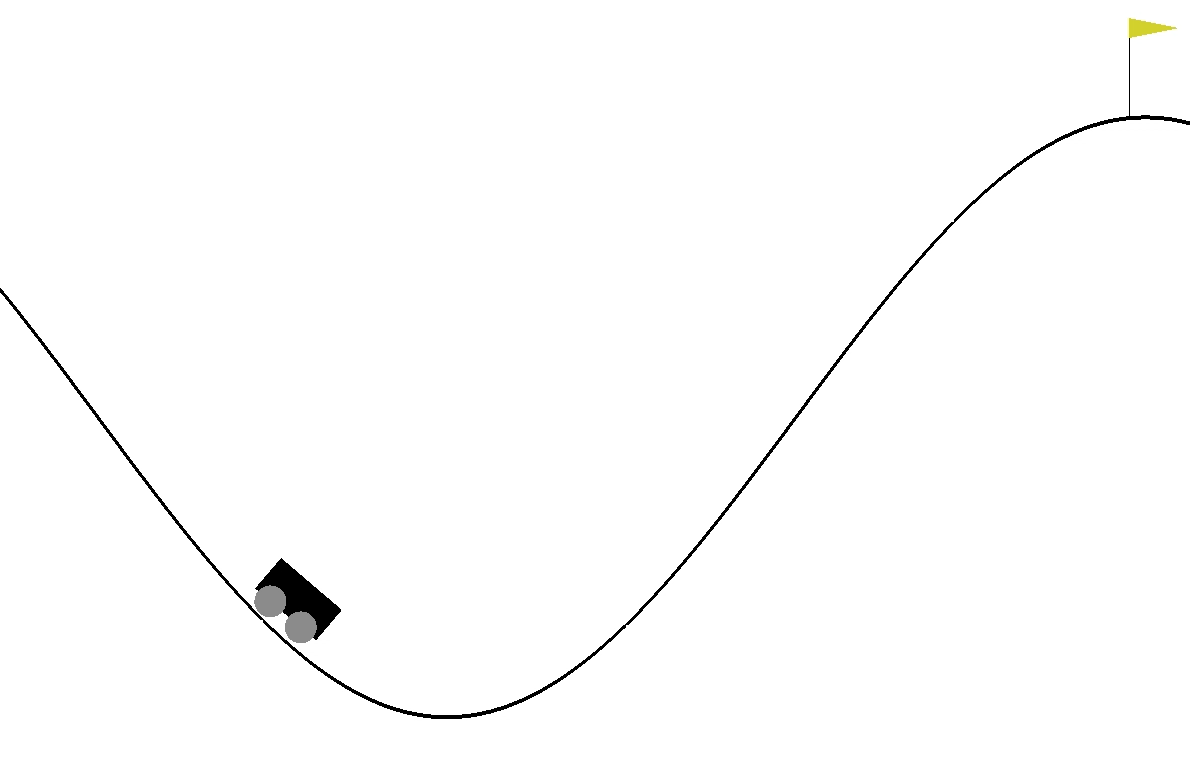}}%
\end{minipage}
\centering
\begin{minipage}{0.32\columnwidth}
\centering
\subfigure[Contours of $f(x)$]{\label{fig:mountain-b}\includegraphics[width=0.9\textwidth]{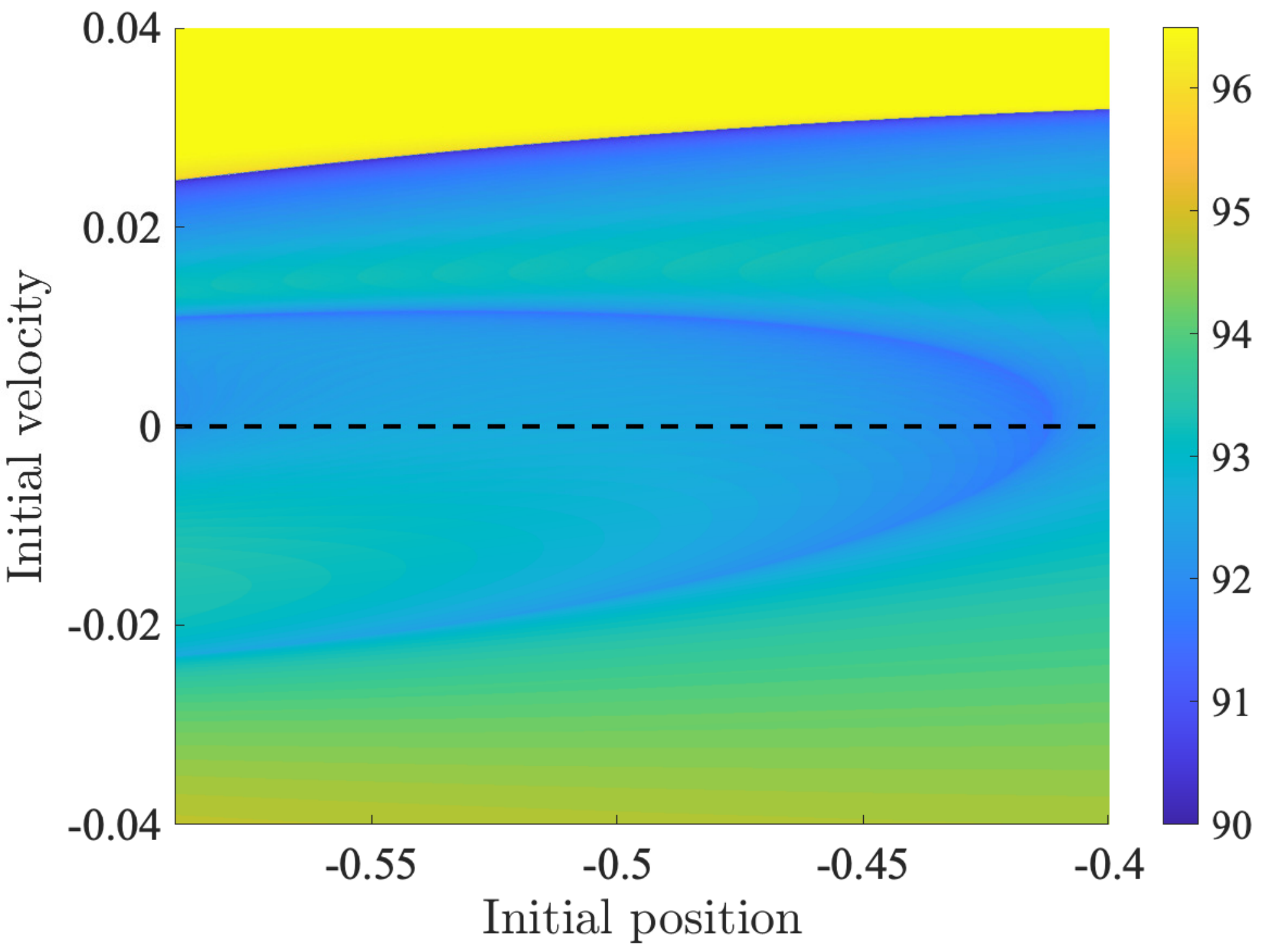}}%
\end{minipage}
\begin{minipage}{0.32\columnwidth}
\centering
\subfigure[Ratio of variance vs. $p_{\gamma_{\mathrm{test}}}$]{\label{fig:mountain-c}\includegraphics[width=0.9\textwidth]{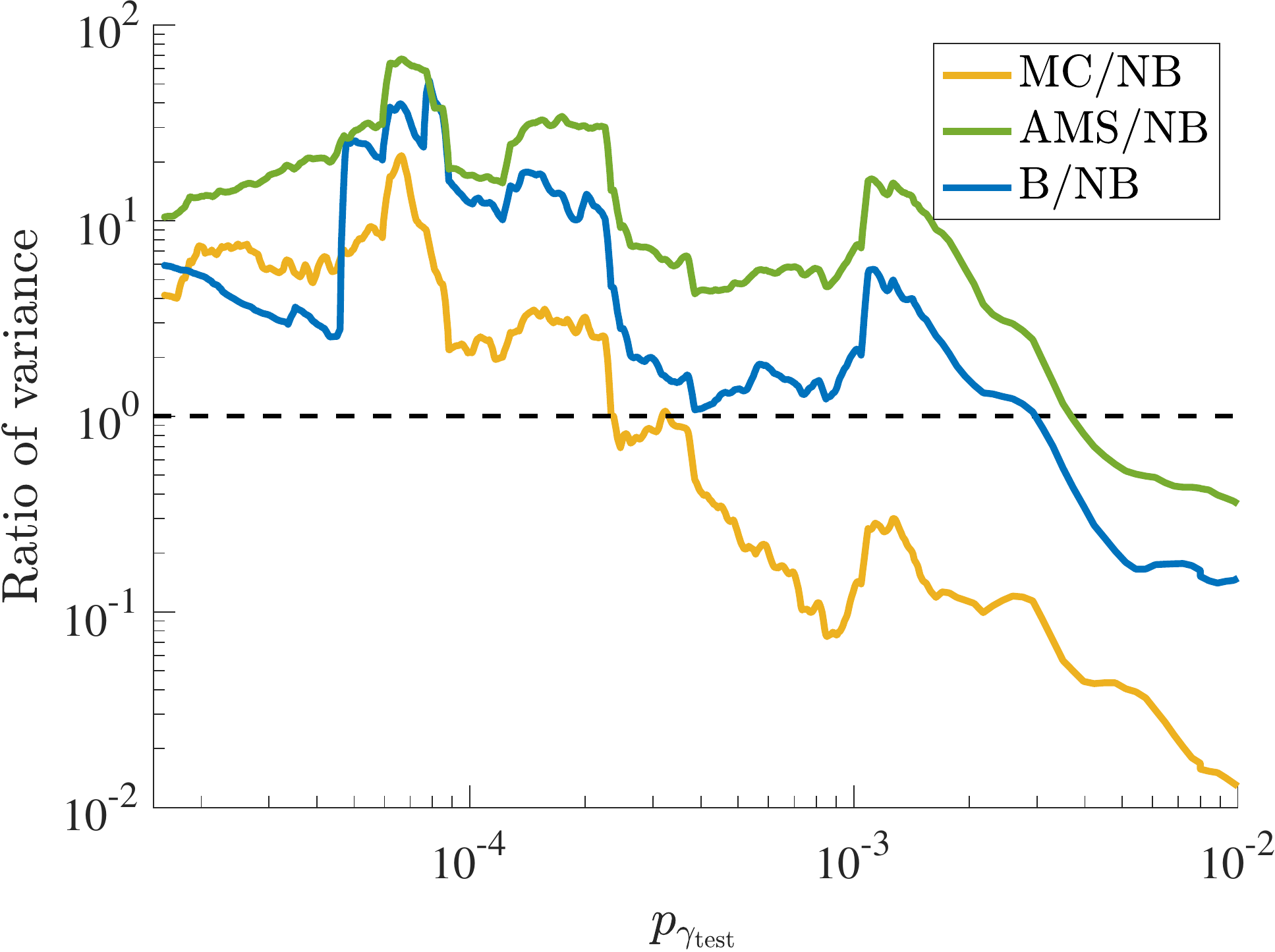}}%
\end{minipage}
\caption[]{\label{fig:mountain}Experiments on the MountainCar environment. The dashed horizontal line in (b) is the line along which the controller is formally verified. 10 trials are used for the variance ratios in (c). The irregular geometry degrades performance of AMS and B, but B benefits slightly from gradients over AMS. NB uses gradients and neural warping to outperform all other techniques.}
\ifarxiv
\else
\vskip -10pt
\fi
\end{figure}

\paragraph{Sensitivity of a formally-verified controller under domain perturbation}
We consider a minimal reinforcement learning task, the MountainCar problem \cite{moore1990efficient} (Figure \ref{fig:mountain-a}).~\citet{ivanov2019verisig} created a formally-verified neural network controller to achieve reward $> 90$ over all initial positions $\in [-0.59, -0.4]$ and 0 initial velocity (see Appendix \ref{app:appendix_experiments}). The guarantees of formal verification hold only with respect to the specified domain; even small domain perturbations can affect system performance~\cite{ivanov2020case}. We illustrate this sensitivity by adding a small perturbation to the initial velocity $\sim \mathcal{N}(0, 10^{-4})$ and seek $p_{\gamma}:=\P_0(\mathrm{reward} \le 90)$ for $P_0\!=\!\mathrm{Unif}(-0.59, -0.4)\! \times\!\mathcal{N}(0, 10^{-4})$. We measure the ground-truth failure rate as $p_{\gamma}=1.6 \cdot 10^{-5}$ using 50 million naive Monte Carlo samples.

Figure \ref{fig:mountain-b} shows contours of $f(x)$. Notably, the failure region (dark blue) is an extremely irregular geometry with pathological curvature, which renders MCMC difficult for AMS and B \cite{betancourt2017conceptual}. Quantitatively, poor mixing adversely affects the performance of AMS and B, and they perform even worse than MC (Figure \ref{fig:mountain-c}). Whereas gradients help B slightly over AMS, gradients and neural warping together help NB outperform all other methods. We next move to higher-dimensional systems.

\begin{figure}[t]
\begin{minipage}{0.32\columnwidth}
\centering
\ifarxiv
	\subfigure[Rocket landing]{\label{fig:rocket-a}\includegraphics[width=0.75\textwidth]{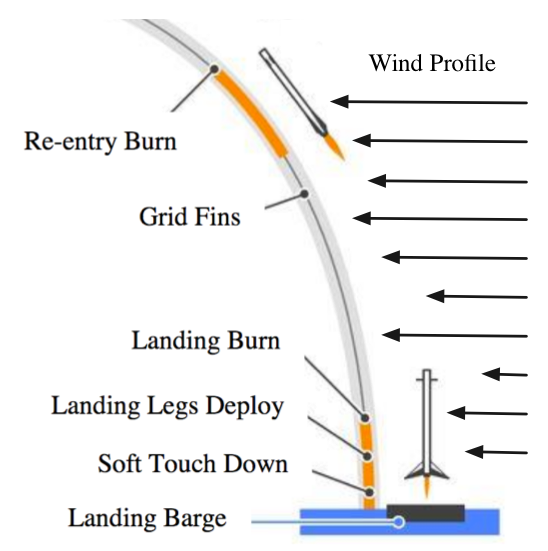}}%
\else
	\subfigure[Rocket landing]{\label{fig:rocket-a}\includegraphics[height=90pt,width=97pt]{./figs/rocket.png}}%
\fi
\end{minipage}
\centering
\begin{minipage}{0.32\columnwidth}
\centering
\ifarxiv
	\subfigure[Failure rates]{\label{fig:rocket-b}\includegraphics[width=1.0\textwidth]{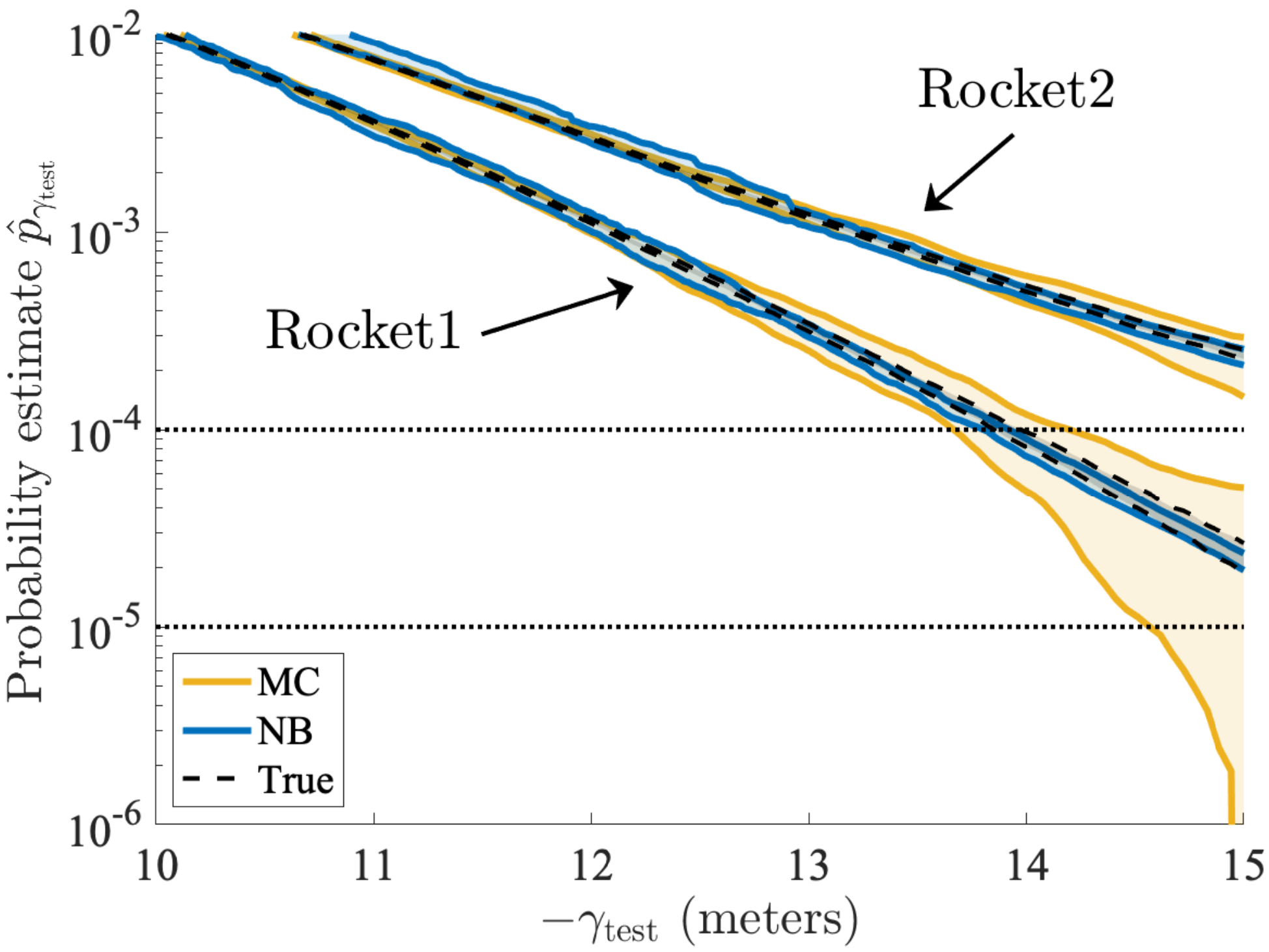}}%
\else
	\subfigure[Failure rates]{\label{fig:rocket-b}\includegraphics[height=90pt]{./figs/rocket_probs}}%
\fi
\end{minipage}
\begin{minipage}{0.32\columnwidth}
\centering
\ifarxiv
	\subfigure[Failure modes]{\label{fig:rocket-c}\includegraphics[width=1.0\textwidth]{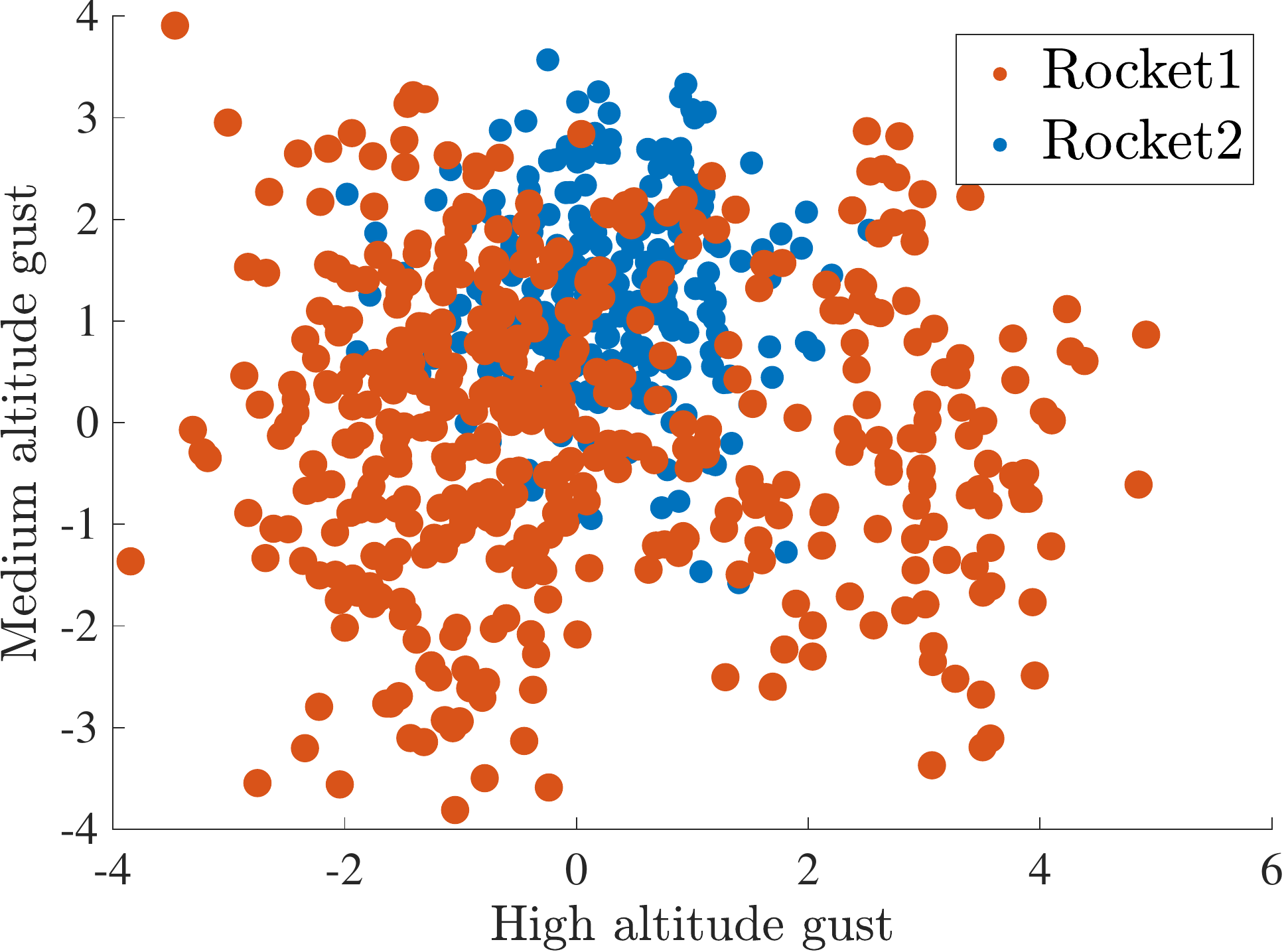}}%
\else
	\subfigure[Failure modes]{\label{fig:rocket-c}\includegraphics[height=90pt]{./figs/rocket_modes}}%
\fi
\end{minipage}
\caption[]{\label{fig:rocket}Rocket design experiments. NB's high-confidence estimates enable quick design iterations to either increase the landing pad radius or consider a third rocket that fails with probability $<10^{-5}$. Low-dimensional visualization shows that Rocket2's failure types are more concentrated than those of Rocket1, even though Rocket2 has a higher overall probability of failure.}
	\ifarxiv
\else
\vskip -20pt
\fi
\end{figure}

\paragraph{Rocket design}
We now consider the problem of autonomous, high-precision vertical landing of an orbital-class rocket (Figure \ref{fig:rocket-a}), a technology first demonstrated by SpaceX in 2015. Rigorous system-evaluation techniques such as our risk-based framework are powerful tools for quickly exploring design tradeoffs. In this experiment, the amount of thrust which the rocket is capable of deploying to land safely must be balanced against the payload it is able to carry to space; stronger thrust increases safety but decreases payloads. We consider two rocket designs and we evaluate their respective probabilities of failure (not landing safely on the landing pad) for landing pad sizes up to $15$ meters in radius. That is, $-f(x)$ is the distance from the landing pad's center at touchdown and $\gamma=-15$. We evaluate whether the rockets perform better than a threshold failure rate of $10^{-5}$.

We let $P_0$ be the 100-dimensional search space parametrizing the sequence of wind-gusts during the rocket's flight. Appendix \ref{app:appendix_experiments} contains details for this parametrization and the closed-loop simulation of the rocket's control law (based on industry-standard approaches~\cite{blackmore2017autonomous, ridderhof2019minimum}). Figure \ref{fig:rocket-b} shows the estimated performance of the two rockets. We show only MC and NB for clarity; comparisons with other methods are in Table \ref{tab:results} (with ground-truth values calculated using 50 million naive Monte Carlo simulations). Whereas both NB and MC confidently estimate Rocket2's failure rate as higher than $10^{-4}$, only NB confidently estimates Rocket1's failure rate as higher than $10^{-5}$, letting engineers quickly judge whether to increase the size of the landing pad or build a better rocket. 

We can also distinguish between the modes of failure for the rockets. Namely, Figure \ref{fig:rocket-c} shows a PCA projection of failures (with $\gamma_{\rm test}=-15$) onto 2 dimensions. Analysis of the PCA modes indicates that failures are dominated by high altitude and medium altitude gusts. Even though Rocket2 has a higher probability of failure, its failure mode is more concentrated than Rocket1's failures.

\begin{figure}[!!t]
\begin{minipage}{0.32\columnwidth}
\centering
\ifarxiv
	\subfigure[The environment]{\label{fig:carracing-a}\includegraphics[height=0.7\textwidth,width=1.0\textwidth]{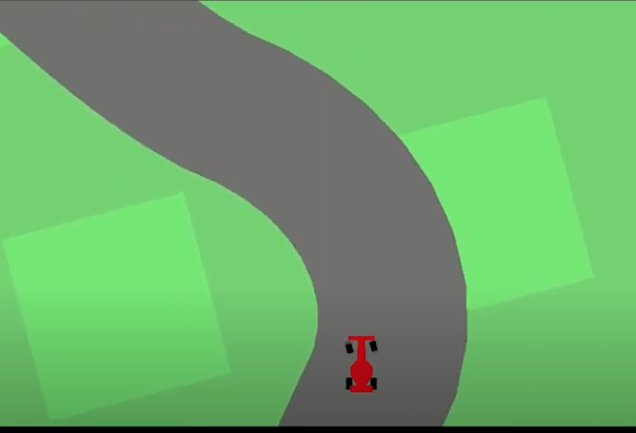}}%
\else
	\subfigure[The environment]{\label{fig:carracing-a}\includegraphics[height=90pt,width=110pt]{./figs/carracing_env.png}}%
\fi
\end{minipage}
\centering
\begin{minipage}{0.32\columnwidth}
\centering
\ifarxiv
	\subfigure[Failure rates]{\label{fig:carracing-b}\includegraphics[width=1.0\textwidth]{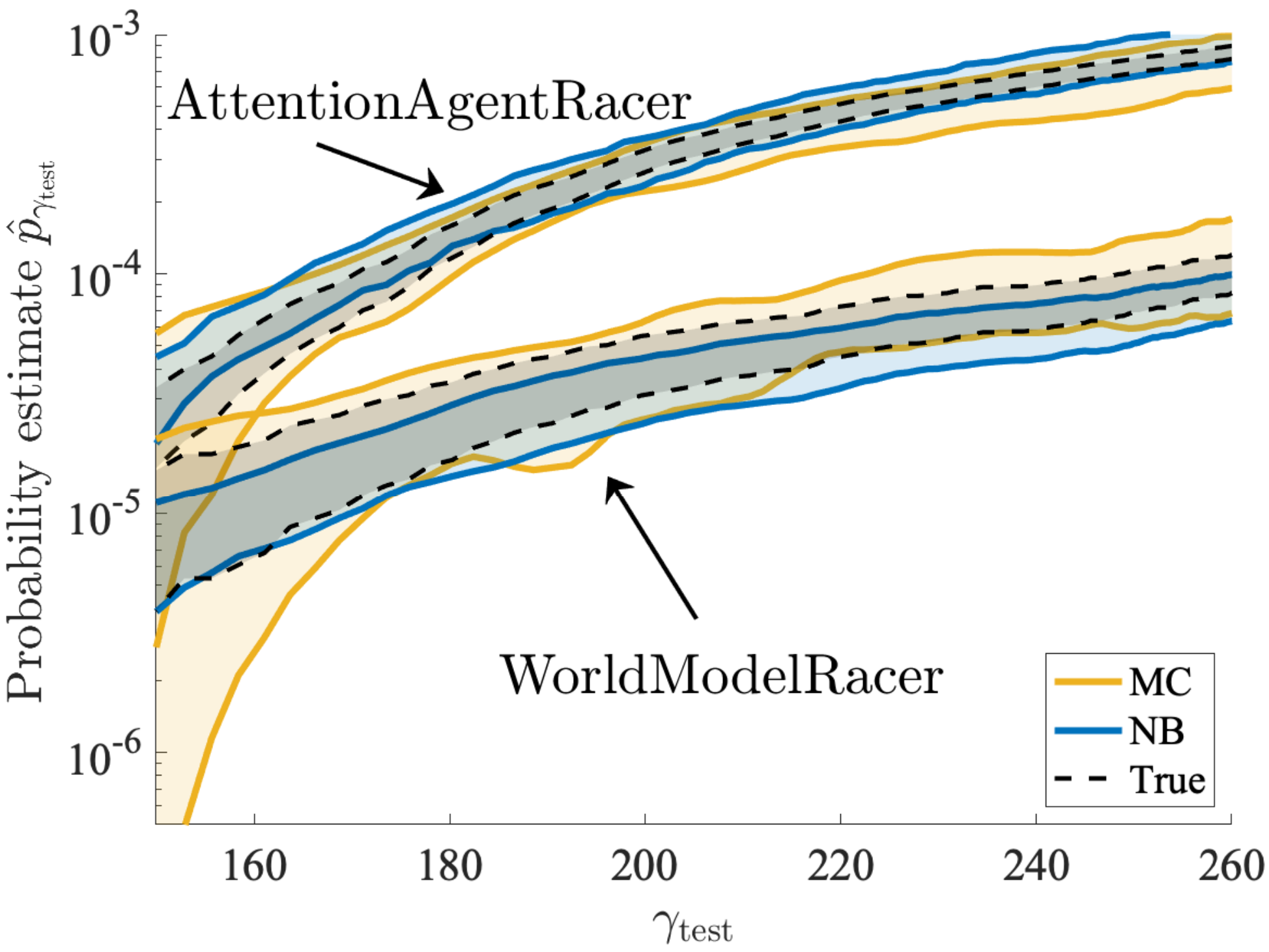}}%
\else
	\subfigure[Failure rates]{\label{fig:carracing-b}\includegraphics[height=90pt]{./figs/carracing_probs}}%
\fi
\end{minipage}
\begin{minipage}{0.32\columnwidth}
\centering
\ifarxiv
	\subfigure[Failure modes]{\label{fig:carracing-c}\includegraphics[width=1.0\textwidth]{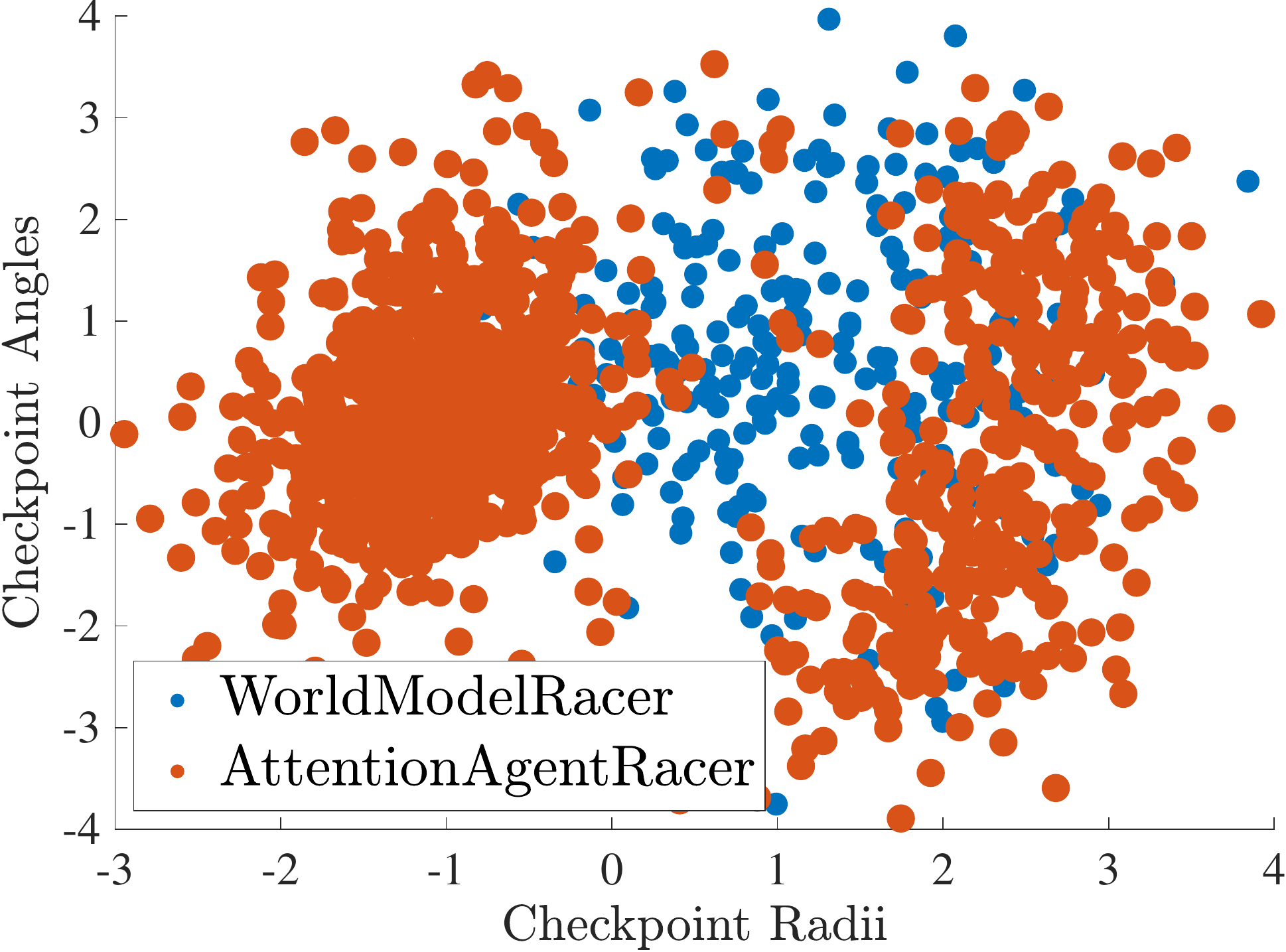}}%
\else
	\subfigure[Failure modes]{\label{fig:carracing-c}\includegraphics[height=90pt]{./figs/car_modes}}%
\fi
\end{minipage}
\caption[]{\label{fig:carracing}CarRacing experiments. MC cannot distinguish between the policies below $\gamma_{\mathrm{test}}=160$.  NB's high-confidence estimates enable model comparisons at extreme limits of failure. Low-dimensional visualization of the failure modes shows that the algorithms fail in distinct ways.}
\ifarxiv
\else
\vskip -10pt
\fi
\end{figure}

\paragraph{Car racing}
The CarRacing environment (Figure \ref{fig:carracing-a}) is a challenging reinforcement-learning task with a continuous action space and pixel observations.  Similar observation spaces have been proposed for real autonomous vehicles (\eg~\cite{bansal2018chauffeurnet,luo2018fast,wang2019monocular}). We compare two recent approaches, AttentionAgentRacer \cite{tang2020neuroevolution} and WorldModelRacer \cite{ha2018world} that have similar average performance: they achieve average rewards of $903\pm 49$ and $899\pm 46$ respectively (mean $\pm$ standard deviation over 2 million trials). Both systems utilize one or more deep neural networks to plan in image-space, so neither has performance guarantees.
We evaluate the probability of getting small rewards ($\gamma= 150$). 

The 24-dimensional search space $P_0$ parametrizes the generation of the racing track (details are in Appendix \ref{app:appendix_experiments}). This environment does not easily provide gradients due to presence of a rendering engine in the simulation loop. Instead, we fit a Gaussian process surrogate model to compute $\nabla f(x)$ (see Appendix \ref{app:appendix_experiments}). As these experiments are extremely expensive (taking up to 1 minute per simulation), we only use 2 million naive Monte Carlo samples to compute the ground-truth failure rates. Figure \ref{fig:carracing-b} shows that, even though the two models have very similar average performance, their catastrophic failure curves are distinct. Furthermore, MC is unable to distinguish between the policies below rewards of 160 due to its high uncertainty, whereas NB clearly shows that WorldModelRacer is superior. Note that, because even the ground-truth has non-negligible uncertainty with 2 million samples, we only report the variance component of relative mean-square error in Table \ref{tab:results}.

As with the rocket design experiments, we visualize the modes of failure (defined by $\gamma_{\rm test} = 225$) via PCA in Figure \ref{fig:carracing-c}. The dominant eigenvectors involve large differentials between radii and angles of consecutive checkpoints that are used to generate the racing tracks. AttentionAgentRacer has two distinct modes of failure, whereas WorldModelRacer has a single mode.
\ifarxiv
\else
\vskip -5pt
\fi

\ifarxiv
\begin{table}[t]\centering\begin{scriptsize}
\else
\begin{table}[!!t]\centering\begin{scriptsize}
\fi
  \caption{Relative mean-square error $\E[\left ({\hat p_{\gamma}}/{p_{\gamma}}-1\right )^2]$ over 10 trials }
  \centering
  \begin{tabular}{lllllll}
    \toprule\label{tab:results}
            & Synthetic  & MountainCar & Rocket1 & Rocket2 & AttentionAgentRacer & WorldModelRacer \\
    \midrule
    MC      & 1.1821     & 0.2410      & 1.1039  & 0.0865  & 1.0866     & 0.9508  \\
    AMS     & 0.0162     & 0.5424      & 0.0325  & 0.0151  & 1.0211     & 0.8177  \\
    B       & 0.0514     & 0.3856      & 0.0129  & 0.0323  & 0.9030     & 0.7837  \\
    NB      & \textbf{0.0051}     & \textbf{0.0945}      & \textbf{0.0102}  & \textbf{0.0078}  & \textbf{0.2285}     & \textbf{0.1218}  \\
    \midrule
    $p_{\gamma}$ & $3.6\cdot 10^{-6}$ &$1.6 \cdot 10^{-5}$ & $2.3\cdot10^{-5}$ & $2.4\cdot10^{-4}$ & $\approx2.5\cdot 10^{-5}$ & $\approx9.5 \cdot 10 ^{-6}$\\
    \bottomrule
  \end{tabular}\end{scriptsize}
  	\ifarxiv
	\else
	\vskip -10pt
	\fi
\end{table}

\section{Conclusion}
There is a growing need for rigorous evaluation of safety-critical systems which contain components without formal guarantees (\eg~deep neural networks). Scalably evaluating the safety of such systems in the presence of rare, catastrophic events is a necessary component in enabling the development of trustworthy high-performance systems. Our proposed method, neural bridge sampling, employs three concepts---exploration, exploitation, and optimization---in order to evaluate system safety with provable statistical and computational efficiency.
We demonstrate the performance of our method on a variety of reinforcement-learning and robotic systems, highlighting its use as a tool for continuous integration and rapid engineering design. In future work, we intend to investigate how efficiently sampling rare failures---like we propose here for \emph{evaluation}---could also enable the \emph{automated repair} of safety-critical reinforcement-learning agents.

\section*{Broader Impact}
This paper presents both foundational theory and methods for efficiently evaluating the performance of safety-critical autonomous systems. By definition, such systems can cause injury or death if they malfunction~\cite{bowen1993safety}. Thus, improving the tools that practitioners have to perform risk-estimation has the potential to provide a strong positive impact. On the other hand, the improved scalability of our method could be used to more efficiently find (zero-day) exploits and failure modes in $P_0$ (the model of the operational design domain). However, we note that adversarial examples or exploits can also be found via a variety of purely optimization-based methods~\cite{akhtar2018threat}. The nuances of our method are primarily concerned with the frequency of adverse events, an extra burden; thus, we anticipate they will be of little interest to malicious actors who can manipulate the observations and sensor measurements of complex systems. Another potential concern about the use of our method is with respect to the identification of $P_0$, which we specifically assume to be known in this paper. The gap between $P_0$ in simulation and the real distribution of the environment could lead to overconfidence in the capabilities of the system under test. In Section~\ref{sec:related} we outline complementary work in anomaly detection and distributionally robust optimization which could mitigate such risks. Still, more work needs to be done to standardize the operational domain of specific tasks by regulators and technology-stakeholders. Nevertheless, we believe that our method will enable the comparison of autonomous systems in a common language---risk---across the spectrum from engineers to regulators and the public. 

The applications of our technology are diverse (\cf~\citet{corso2020survey}), ranging from testing autonomous vehicles~\cite{okelly2018scalable, norden2019efficient} and medical devices~\cite{o2018silico} to evaluating deep neural networks~\cite{webb2018statistical} and reinforcement-learning agents~\cite{uesato2018rigorous}. In the case of autonomous vehicles,~\citet{sparrow2017human} argue that it will be morally wrong not to deploy self-driving technology once performance exceeds human capabilities. Our work is an important tool for determining when this performance threshold is achieved due to the rare nature of serious accidents~\citep{kalra2017challenges}. 
While the widespread availability of autonomy-enabled devices could narrowly benefit public health, there are many external risks associated with their development. First, many learning-based components of these systems will require massive and potentially invasive data collection~\cite{re2018software}; preserving privacy of the public via federated learning~\cite{mcmahan2017communication} and differential privacy-based mechanisms~\cite{dwork2008differential} should remain important initiatives within the machine-learning community. A second potential negative consequence of the applications like autonomous vehicles is the use of the real-world as a ``simulator'' within a reinforcement-learning scheme by releasing ``beta'' autonomy features (\eg~Tesla Autopilot~\cite{karpathy2017software}). Unlike established industries such as aerospace~\cite{tosney2015space}, many potential applications currently lack regulation and standards; it is important to ensure that industry works with policy makers to develop safety standards in a way that avoids regulatory capture. 
If widely adopted in regulatory frameworks, our tool would enable rational decisions about the impact, positive or negative, of safety-critical autonomous systems before real lives are affected. 

More broadly, the advent of autonomy could spark significant societal changes. For example, the autonomous applications described previously could become core components of weapons systems and military technology that are incompatible with (modern interpretations of) just war theory~\cite{sparrow2007killer}. Similarly, the automation of the transportation industry has the potential to rapidly destroy the economics of public infrastructure and cost millions of jobs~\cite{sparrow2017human}. 
Thus, \citet{benkler2019don} highlights that there is a growing need for the academic community to take action on defining the broader performance criteria to which we will hold AI applications.~\citet{brundage2020toward} and \citet{wing2020trustworthy} outline broad research agendas which are necessarily interdisciplinary. Still, much more work needs to be done to empower researchers to influence policy. These efforts will require systemic initiatives by research institutions and organizations to engage with local, national, and international governing bodies.

\section*{Acknowledgements}
AS and JD were partially supported by the DAWN Consortium, NSF CAREER CCF-1553086, NSF HDR 1934578 (Stanford Data Science Collaboratory), ONR YIP N00014-19-2288, and the Sloan Foundation. MOK was supported by an NSF GRFP Fellowship. RT was supported by Lincoln Laboratory/Air Force Award No. PO\# 7000470769 and Amazon Robotics Award No. CC MISC 00272683 2020 TR. 

\setlength{\bibsep}{3pt}
\bibliographystyle{abbrvnat}

\newpage

\appendix

\section{Warped Hamiltonian Monte Carlo (HMC)}
\label{app:hmc}

\begin{algorithm}[t]
  \caption{\label{alg:warpedhmc} WarpedHMC}
  \begin{small}
  \begin{algorithmic}[]
    \State \textbf{Input:} Sample $x$, momentum $v \sim \mc N(0,I)$, transform $V_{\theta}$ and its inverse $W_{\theta}$, scale factor $\beta$, step size $\epsilon$
    \State $y \gets W_{\theta}(x)$
    \State $\hat v \gets v -0.5\epsilon\beta I\{f(x) > \gamma\} J_{V_{\theta}}(y)\nabla f(x)$
    \State $\hat y \gets y \cos(\epsilon) + \hat v\sin(\epsilon)$
    \State $\hat v \gets \hat v \cos(\epsilon) - y\sin(\epsilon)$
    \State $\hat x \gets V_{\theta}(\hat y)$
    \State $\hat v \gets \hat v - 0.5\epsilon\beta I\{f(\hat x) > \gamma\} J_{V_{\theta}}(\hat y)\nabla f(\hat x)$
    \State $\hat v \gets -\hat v$
    \State $x \gets \hat x$ with probability $\min(1, \exp(-H(\hat y, \hat v) + H(y, v)))$
    \State  \textbf{Return} $x$
  \end{algorithmic}\end{small}
\end{algorithm}

In this section, we provide a brief overview of HMC as well as the specific rendition, split HMC \cite{shahbaba2014split}. Given ``position'' variables $x$ and ``momentum'' variables $v$, we define the Hamiltonian for a dynamical system as $H(x,v)$ which can usually be written as $U(x) + K(v)$, where $U(x)$ is the potential energy and $K(v)$ is the kinetic energy. For MCMC applications, $U(x)=-\log (\rho_0(x))$ and we take $v \sim \mc N (0, I)$ so that $K(v)=\|v\|^2/2$. In HMC, we start at state $x_i$ and sample $v_i \sim \mc N(0, I)$. We then simulate the Hamiltonian, which is given by the partial differential equations:
\begin{equation*}
\dot x = \frac{\partial H}{\partial v}, \;\;\;\; 	\dot v = -\frac{\partial H}{\partial x}.
\end{equation*}
Of course, this must be done in discrete time for most Hamiltonians that are not perfectly integrable. One notable exception is when $x$ is Gaussian, in which case the dynamical system corresponds to the evolution of a simple harmonic oscillator (\ie~a spring-mass system). When done in discrete time, a symplectic integrator must be used to ensure high accuracy. After performing some discrete steps of the system (resulting in the state $(x_f, v_f)$), we negate the resulting momentum (to make the resulting proposal reversible), and then accept the state $(x_f, -v_f)$ using the standard Metropolis-Hastings criterion: $\min(1, \exp(-H(x_f, -v_f) + H(x_i, v_i)))$ \cite{hastings1970monte}.

The standard symplectic integrator---the leap-frog integrator---can be derived using the following symmetric decomposition of the Hamiltonian (performing a symmetric decomposition retains the reversibility of the dynamics): $H(x, v)= U(x)/2 + K(v) + U(x)/2$. Using simple Euler integration for each term individually results in the following leap-frog step of step-size $\epsilon$:
\begin{align*}
v_{1/2}&=v_i - \frac{\epsilon}{2}\frac{\partial U(x_i)}{\partial x}\\
x_{f}&= x_i + \epsilon \frac{\partial K(v_{1/2})}{\partial v}\\
v_f&= v_{1/2} - \frac{\epsilon}{2}\frac{\partial U(x_f)}{\partial x},
\end{align*}
where each step simply simulates the individual Hamiltonian $H_1(x,v)=U(x)/2$, $H_2(x,v)=K(v)$, or $H_3(x,v)=U(x)/2$ in sequence. As presented by \citet{shahbaba2014split}, this same decomposition can be done in the presence of more complicated Hamiltonians. In particular, consider the Hamiltonian $H(x,v)=U_1(x) + U_0(x) + K(v)$. We can decompose this in the following manner: $H_1(x,v)=U_1(x)/2$, $H_2(x,v)=U_0(x)+K(v)$, and $H_3(x,v)=U_1(x)/2$. We can apply Euler integration to the momentum $v$ for the first and third Hamiltonians and the standard leap-frog step to the second Hamiltonian (or even analytic integration if possible). For this paper, we have $U_0(x) =-\log \rho_0(x)$ and $U_1(x)= -\beta[\gamma -f(x)]_-$. 

To account for warping, the modifications needed to the HMC steps above are simple. When performing warping, we simply perform HMC for a Hamiltonian $\hat H(y,v)$ that is defined with respect to the warped position variable $y$, where $x=V_{\theta}(y)$ for given parameters $\theta$. By construction of the normalizing flows, we assume $y\sim \mc N(0,I)$, so that we can perform the dynamics for $\hat H_2(y,v)$ analytically. Furthermore, the Jacobian $J_{V_{\theta}}(y)$ is necessary for performing the Euler integration of $H_1(y,v)$ and $H_3(y,v)$. This is summarized in Algorithm \ref{alg:warpedhmc}. Note that we always perform the Metropolis-Hastings acceptance with respect to the true Hamiltonian $H$, rather than the Hamiltonian $\hat H$ that assumes perfect training of the normalizing flows. 

\paragraph*{HMC and non-smooth functions}
In Section \ref{sec:approach}, we assumed that the measure of non-differentiable points is zero for the energy potentials considered by HMC. As discussed by \citet{afshar2015reflection}, the inclusion of the Metropolis-Hastings acceptance criterion as well as the above assumption ensures that HMC asymptotically samples from the correct distribution even for non-smooth potentials. An equivalent intuitive explanation for this can be seen by viewing the ReLU function $[x]_{+}$ as the limit of softplus functions $g_k(x):=\log(1+\exp(kx))/k$ as the sharpness parameter $k \to \infty$. We can freely choose $k$ such that, up to numerical precision, Algorithm \ref{alg:warpedhmc} is the same whether we consider using a ReLU or sufficiently sharp (\eg~large $k$) softplus potential, because, with probability one, we will not encounter the points where the potentials differ. When further knowledge about the structure of the non-differentiability is known, the acceptance rate of HMC proposals can be improved \cite{pakman2013auxiliary,lan2014spherical,pakman2014exact,afshar2015reflection,chaari2016hamiltonian}.

\section{Performance analysis}
\label{app:performance}
\subsection{Proof of Proposition \ref{prop:main}}

We begin with showing the convergence of the number of iterations. To do this, we first show almost sure convergence of $\beta_k$ in the limit $N \to \infty$. We note that in the optimization problem~\eqref{eq:beta_anneal}, $\beta_{k}$ is a feasible point, yielding $b_k(\beta)=1$. Thus, $\beta_{k+1}\ge \beta_{k}\ge \beta_0:=0$. Due to this growth of $\beta_k$ with $k$, we have
\begin{equation*}
\frac{Z_{k+1}}{Z_k} = \E_{P_k} \left [\frac{\rho_{k+1}(X)}{\rho_k(X)} \right ]	\le 1,
\end{equation*}
\begin{align*}
	\P_k(\obj(X) \le \thresh) &= \E_{P_{k+1}}\left [ \frac{Z_{k+1}}{Z_k} \frac{\rho_{k} (X)}{\rho_{k+1}(X)}I\{ f(X) \le \gamma)\} \right ]\\
	&=\frac{Z_{k+1}}{Z_k}\E_{P_{k+1}}\left [ I\{ f(X) \le \gamma)\} \right ] \\
	&\le \P_{k+1}(\obj(X) \le \thresh).
\end{align*}
By the unfiorm convergence of empirical measures offered by the Glivenko-Cantelli Theorem, the value $a_k \to \P_{k}(f(X) \le \gamma)$ almost surely. Then, the stop condition can be rewritten as $b_k(\beta) \ge a_k/s \to \P_{k}(f(X) \le \gamma)/s \ge p_{\gamma}/s$. Since $b_k(\beta)$ is monotonically decreasing in the quantity $\beta-\beta_k$, this constraint gives an upper bound for $\beta_{k+1}$, and, as a result, all $\beta_k$ are almost surely bounded from above and below. We denote this interval as $\mc{B}$. 

Now, we consider the convergence of the solutions to the finite $N$ versions of problem~\eqref{eq:beta_anneal}, denoted $\beta_k^N$, to the ``true'' optimizers $\beta_k$ in the limit as $N\to \infty$. Leaving the dependence on $\beta_k$ implicit for the moment, we consider the random variable $Y:= g(X;\beta):=\exp \left ( (\beta - \beta_k)[\gamma - f(X)]_{-} \right )$. Then, since $\beta \in \mc{B}$ is bounded and $g$ is continuous in $\beta$, we can state the Glivenko-Cantelli convergence of the empirical measure uniformly over $\mc{B}$: $\sup_{\beta \in \mc{B}} \| F^N(Y) - F(Y) \|_{\infty} \to 0$ almost surely, where $F$ is the cumulative distribution function for $Y$. Note that the constraints in the problem~\eqref{eq:beta_anneal} can be rewritten as expectations of this random variable $Y$. Furthermore, the function $g$ is strictly monotonic in $\beta$ (and therefore invertible) for non-degenerate $f(X)$ (\ie~$f(x) > \gamma$ for some non-negligible measure under $P_0$). Thus, we have almost sure convergence of the argmin $\beta_{k+1}^N$ to $\beta_{k+1}$.

Until now, we have taken dependence on $\beta_k$ implicitly. Now we make the dependence explicit to show the final step of convergence. In particular, we can write $\beta_{k+1}$ as a function of $\beta_{k}$ (along with their empirical counterparts), For concreteness, we consider the following decomposition for two iterations:
\begin{equation*}
	|\beta_{2}^N(\beta_1^N) - \beta_{2}(\beta_1)| \le | \beta_{2}^N(\beta_1^N) - \beta_{2}(\beta_1^N) | + | \beta_{2}(\beta_1^N) - \beta_{2}(\beta_1) |.
\end{equation*}
We have already shown above that the first term on the right hand side vanishes almost surely. By the same reasoning, we know that $\beta_1^N \to \beta_1$ almost surely. The second term also vanishes almost surely since $\beta_{k+1}(\beta)$ is a continuous mapping. This is due to the fact that the constraint functions in problem~\eqref{eq:beta_anneal} are continuous functions of both $\beta$ and $\beta_k$ along with the invertibility properties discussed previously. Then, we simply extend the telescoping series above for any $k$ and similarly show that all terms vanish almost surely. This shows the almost sure convergence for all $\beta_k$ up to some $K$.

Now we must show that $K$ is bounded and almost surely converges to a constant. To do this we explore the effects of the optimization procedure. Assuming the stop condition (the second constraint) does not activate, the first constraint in problem~\eqref{eq:beta_anneal} has the effect of making $\Z_{k+1}/Z_{k}=\alpha$ (almost surely), which implies $\P_{k+1}(\obj(X) \le \thresh) = \P_{k}(\obj(X) \le \thresh)/\alpha$. In other words, we magnify the event of interest by a factor of $1/\alpha$. The second constraint can be rewritten as $\P_{k+1}(f(X) \le \gamma) \le s$. Thus, we magnify the probability of the region of interest by factors of $\alpha$ unless doing so would increase the probability to greater than $s$. In that case, we conclude with setting the probability to $s$ (since $\P_\beta(f(X) \le \gamma)$ is monotonically increasing in $\beta$). In this way, we have 0 iterations for $p_{\gamma} \in [s, 1]$, 1 iteration for $p_{\gamma} \in [\alpha s, s)$, 2 iterations for $p_{\gamma} \in [\alpha^2 s, \alpha s)$, and so on. Then, the total number of iterations is (almost surely) $\floor{ \log(p_\gamma) / \log(\alpha) } + I\{p_{\gamma}/\alpha^{\floor{ \log(p_\gamma) / \log(\alpha) }} < s \}$.

Now we move to the relative mean-square error of $\hat p_{\gamma}$. We employ the delta method, whereby, for large $N$, this is equivalent to $\text{Var}(\log (\hat p_{\gamma}))$ (up to terms $o(1/N)$). For notational convenience, we decompose $\widehat E_k$ into its numerator and denominator:
\begin{align*}
  A_k(X):=\rho^B_k (X)/\rho_{k-1}(X),&\qquad \widehat A_k:= \frac{1}{N}\sum_{i=1}^N A_k(x_i^{k-1})\\
    B_k(X):=\rho^B_k (X)/\rho_{k}(X),  &\qquad  \widehat B_k:= \frac{1}{N}\sum_{i=1}^N B_k(x_i^k).
\end{align*}
By construction (and assumption of large $T$), Algorithm \ref{alg:main} has a Markov property that each iteration's samples $x_i^k$ are independent of the previous iterations' samples $x_i^{k-1}$ given $\beta_k$. For shorthand, let $\beta_{0:k}$ denote all $\beta_0, \ldots, \beta_k$. Conditioning on $\beta_{0:k}$, we have
\begin{align*}
\text{Var}(A_k) &= \text{Var}\left ( \E [A_k | \beta_{0:k}] \right) + \E \left [\text{Var}\left ( A_k | \beta_{0:k}\right ) \right ].
\end{align*}
Since $\beta_{0:k}$ approaches constants almost surely as $N \to \infty$, the first term vanishes and the second term is the expectation of a constant. In particular, the second term is as follows:
\begin{align*}
\text{Var}\left ( A_k | \beta_{0:k} \right ) & = \E \left [A_k^2 | \beta_{0:k} \right ]  - \left ( \E \left [ A_k|\beta_{0:k} \right ]\right)^2\\
&= \E_{P_{k-1}} \left [  \frac{\rho_k(X)}{\rho_{k-1}(X)} \right]  - \left (  \E_{P_{k-1}} \left [  \sqrt\frac{\rho_k(X)}{\rho_{k-1}(X)} \right]\right)^2\\
&= \frac{Z_k}{Z_{k-1}} - \left ( \frac{Z_k^B}{Z_{k-1}} \right )^2.
\end{align*}
Similarly, $\text{Var}(B_k|\beta_{0:k})= Z_{k-1}/Z_k - (Z_k^B/Z_{k})^2$. Next we look at the covariance terms:
\begin{align*}
\text{Cov}(A_{k-1}, A_{k}) = \text{Cov}\left ( \E [A_{k-1} | \beta_{0:k}], \E [A_{k} | \beta_{0:k}] \right) + \E \left [\text{Cov}\left ( A_{k-1}, A_k | \beta_{0:k} \right ) \right ].
\end{align*}
Again, the first term vanishes since $\beta_{0:k}$ approach constants as $N \to \infty$. By construction, the second term is also 0 since the quantities are conditionally independent. Similarly, $\text{Cov}(B_{k-1}, B_{k})=0$ and $\text{Cov}(A_{i}, B_{j})=0$ for $j \neq i-1$. However, there is a nonzero covariance for the quantities that depend on the same distribution:
\begin{align*}
\text{Cov}\left ( B_{k}, A_{k+1} | \beta_{0:k+1} \right ) & = \E \left [B_k A_{k+1} | \beta_{0:k+1} \right ]  - \E \left [ B_k|\beta_{0:k+1} \right ] \E \left [ A_{k+1}|\beta_{0:k+1} \right ]\\
&= \E_{P_{k}} \left [  \frac{\sqrt {\rho_{k-1}(X)\rho_{k+1}(X)} }{\rho_{k}(X)} \right]  - \frac{Z_{k+1}^B}{Z_{k}}\frac{Z_k^B}{Z_k}\\
&= \frac{Z_k^C}{Z_{k}} - \frac{Z_{k+1}^B}{Z_{k}}\frac{Z_k^B}{Z_k}.
\end{align*}
By the large $T$ assumption, the samples $x_i^k$ and $x_j^k$ are independent for all $i \neq j$ given $\beta_k$. Then we have
\begin{equation*}
	\text{Var}(\widehat A_k | \beta_{0:k}) = \text{Var}(A_k|\beta_{0:k})/N, \;\; \text{Var}(\widehat B_k| \beta_{0:k}) = \text{Var}(B_k | \beta_{0:k})/N,
\end{equation*}
\begin{equation*}
\text{Cov}(\widehat B_k, \widehat A_{k+1} | \beta_{0:k+1}) = \text{Cov}(B_k, A_{k+1} | \beta_{0:k+1})/N.	
\end{equation*}
The last term in $\hat p _\gamma$, $\frac{1}{N}\sum_{i=1}^N \frac{\rho_\infty(x^K_i)}{\rho_K(x^K_i)}$, reduces to a simple Monte Carlo estimate since $\frac{\rho_\infty(X)}{\rho_K(X)}= I\{f(X) \le \gamma\}$. Furthermore, this quantity is independent of all other quantities given $\beta_{0:K}$ and, as noted above, approaches $s$ almost surely as $N\to \infty$. 

Putting this all together, the delta method gives (as $N \to \infty$ so that $\beta_{0:K}$ approach constants almost surely), 
\begin{align*}
	\text{Var}(\log (\hat p_{\gamma})) &\to \sum_{k=1}^K  \left ( \frac{\text{Var}(\widehat A_k)}{ (Z_k^B/Z_{k-1} )^2} + \frac{\text{Var}(\widehat B_k)}{ (Z_k^B/Z_{k})^2 } \right ) -2 \sum_{k=1}^{K-1} \frac{\text{Cov}( \widehat B_k, \widehat A_{k+1})}{  Z_{k+1}^B Z_k^B /Z_{k}^2  } + \frac{1-s}{sN}+ o \left (\frac{1}{N} \right ).
\end{align*}
The Bhattacharrya coefficient can be written as
\begin{equation*}
G(P_{k-1}, P_k)=\int_\mc{X}\sqrt{\frac{\rho_{k-1}(x)}{Z_{k-1}}\frac{\rho_k(x)} {Z_k}}dx = \frac{Z_k^B}{\sqrt{Z_{k-1} Z_k}}.
\end{equation*}
Furthermore, we have
\begin{equation*}
	\frac{G(P_{k-1}, P_{k+1})}{G(P_{k-1}, P_{k})G(P_{k}, P_{k+1})} = \frac{Z_k^C}{\sqrt{Z_{k-1}Z_{k+1}}} \frac{\sqrt{Z_{k-1}Z_k}}{Z_k^B} \frac{\sqrt{Z_{k}Z_{k+1}}}{Z_{k+1}^B}=\frac{Z_k^CZ_k}{Z^B_{k} Z^B_{k+1}},
\end{equation*}
yielding this final result
\begin{scriptsize}
\begin{equation}\label{eq:final}
\text{Var}(\log (\hat p_{\gamma})) \to \frac{2}{N}\sum_{k=1}^K \left (\frac{1}{G(P_{k-1}, P_k)^2}-1 \right ) -\frac{2}{N} \sum_{k=1}^{K-1} \left ( \frac{G(P_{k-1}, P_{k+1})}{G(P_{k-1}, P_{k})G(P_{k}, P_{k+1})} -1\right ) +\frac{1-s}{sN} + o \left (\frac{1}{N} \right ).
\end{equation}\end{scriptsize}\noindent We remark that a special case of this formula is for $K=1$ and $s=1$ (so only the first term survives), which is the relative mean-square error for a single bridge-sampling estimate $\widehat E_k$.

Now, since $G(P, Q) \ge 0$, the terms in the second sum are $\ge -1$ so that the second sum is $\le 2(K-1)/N$. Furthermore, since $s \ge 1/3$, the last term is also $\le 2/N$. Thus, if we have $\frac{1}{G(P_{k-1}, P_k)^2}\le  D $ (with $D \ge 1$), then the asymptotic relative mean-square error \eqref{eq:final} is $\le 2KD/N$ (up to terms $o \left (\frac{1}{N} \right )$).

When performing warping, we follow the exact same pattern as the above results, conditioning on both $\beta_{0:k}$ and $W_{0:k}$,  where $W_0$ is defined as the identity mapping. We follow the same almost-sure convergence proof for $W_k$ as above for $\beta_k$, which requires compactness of $\theta \in \Theta$, continuity of $W$ with respect to $\theta$ and $x$, and that we actually achieve the minimum in problem~\eqref{eq:flowprob}. Although the first two conditions are immediate in most applications, the last condition can be difficult to satisfy for deep neural networks due to the nonconvexity of the optimization problem.

\section{Experimental setups}
\label{app:appendix_experiments}

\subsection{Hyperparameters}
The number of samples $N$ affects the absolute performance of all of the methods tested, but not their relative performance with respect to each other. For all experiments, we use $N=1000$ for B and NB to have adequate absolute performance given our computational budget (see below for the computing architecture used). Other hyperparameters were tuned on the synthetic problem and fixed for the rest of the experiments (with the exception of the MAF architecture for the rocket experiments). The hyperparameters were chosen as follows. 

When performing Hamiltonian dynamics for a Gaussian variable, a time step of $2\pi$ results in no motion and time step of $\pi$ results in a mode reversal, where both the velocity and position are negated. The $\pi$ time step is in this sense the farthest exploration that can occur in phase space (which can be intuitively understood by recognizing that the phase diagram of a simple spring-mass system is a unit circle). Thus, we considered $T=4,8,12,$ and $16$ with time steps $\pi/T$. We found that $T=8$ provided reasonable exploration (as measured by autocorrelations and by the bias of the final estimator $\hat p_{\gamma}$) and higher values of $T$ did not provide much more benefit. For B, we allowed 2 more steps $T=10$ to keep the computational cost the same across B and NB. Similarly, for AMS, we set $T=10$. We also performed tuning online for the time step to keep the accepatance ratio between 0.4 and 0.8. This was done by setting the time step to $\sin^{-1}(\min(1,\sin(t)\exp((p-C)/2))$, where $t$ is the current time step, $p$ is the running acceptance probability for a single chain and $C=0.4$ if $p< 0.4$ or $0.8$ if $p>0.8$. This was done after every $T$ HMC steps.

For the step size of the bridge, we considered $\alpha \in \{0.01, 0.1, 0.3, 0.5\}$. Smaller $\alpha$ results in fewer iterations and better computational efficiency. However, we found that very small $\alpha$ made MAF training difficult (see below for the MAF architectures used). We settled on $\alpha=0.3$, which provided reasonable computational efficiency (no more than 11 iterations for the synthetic problem) as well as stable MAF training. For AMS, we followed the hyperparameter settings of \citet{webb2018statistical}. Namely, we chose a culling fraction of $\alpha_{\mathrm{AMS} }=10\%$, where $\alpha_{\mathrm{AMS}}$ sets the fraction of particles that are removed and rejuvenated at each iteration \cite{webb2018statistical}.

The MAF architectures for the synthetic, MountainCar, and CarRacing experiments were set at 5 MADE units, each with 1 hidden layer of 100 neurons. Because the rocket search space is very high dimensional, we decreased the MAF size for computational efficiency: we set it at 2 MADE units, each with hidden size 400 units. We used 100 epochs for training, a batch size of 100, a learning rate of 0.01 and an exponential learning-rate decay with parameter 0.95.

Given the above parameters, the number of simulations for each experiment varies based on the final probability in question $p_{\gamma}$ (smaller values result in more simulations due to having a higher number of iterations $K$). We had runs of 111000, 101000, 91000, 71000, 91000, and 101000 simulations respectively for the synthetic, MountainCar, Rocket1, Rocket2, AttentionAgentRacer, and WorldModelRacer environments. We used these values as well as the ground truth $p_{\gamma}$ values to determine the number of particles allowed for AMS, $N_{\mathrm{AMS}}=920, 910, 820, 780, 820, 910$ respectively, as AMS has a total cost of $N_{\mathrm{AMS}}(1+\alpha_{\mathrm{AMS}}TK_{\mathrm{AMS}})$, where $K_{\mathrm{AMS}} \approx \log(p_{\gamma})/\log(1-\alpha_{\mathrm{AMS}})$.

For the surrogate Gaussian process regression model for CarRacing, we retrained the model on the most recent $N$ simulations after every $NT$ simulations (\eg~after every $T$ HMC iterations). This made the amortized cost of training the surrogate model negligible compared to performing the simulations themselves. We used a Matern kernel with parameter $\nu=2.5$. We optimized the kernel hyperparameters using an L-BFGS quasi-Newton solver. 

\paragraph*{Computing infrastructure and parallel computation} Experiments were carried out on commodity CPU cloud instances, each with 96 Intel Xeon cores @ 2.00 GHz and 85 GB of RAM. AMS, B, and NB are all designed to work in a Map-Reduce paradigm, where a central server orchestrates many worker jobs followed by synchronization step. AMS requires more iterations and fewer parallel worker threads per iteration than B and NB. In particular, whereas B and NB perform $N$ parallel jobs per iteration, AMS only performs  $\alpha_{\mathrm{AMS}} N_{\mathrm{AMS}}$ parallel jobs per iteration. Thus, B and NB take advantage of massive scale and parallelism much more than AMS.

\subsection{Environment details}

\subsubsection{MountainCar}
The MountainCar environment considers a simple car driving on a mountain road. The car can sense horizontal distance $s$ as well as its velocity $v$, and may send control inputs $u$ (the amount of power applied in either the forward or backward direction). 
The height of the road is given by: $h(s) = 0.45\sin(3s)+0.55$. 
The speed of the car, $v$, is a function of $s$ and $u$ only. Thus, the discrete time dynamics are:	$s_{k+1} = s_k + v_{k+1}$ and $v_{k+1} = v_k + 0.0015u_k -0.0025 \cos(3s_k)$. For a given episode the agent operating the car receives a reward of $-0.1u_k^2$ for each control input and  $100$ for reaching the goal state.

In this experiment we explore the effect of domain shift on a formally verified neural network. We utilize the neural network designed by \citet{ivanov2019verisig}; it contains two hidden layers, each of 16 neurons, for a total of 337 parameters.
For our experiments we use the trained network parameters available at: \url{https://github.com/Verisig/verisig}. \citet{ivanov2019verisig} describe a layer-by-layer approach to verification which over-approximates the reachable set of the combined dynamics of the environment and the neural network. An encoding of this system (network and environment) is developed for the tool Flow$^*$~\cite{chen2013flow} which constructs the (overapproximate) reachable set via a Taylor approximation of the combined dynamics.

The MountainCar environment is considered solved if a policy achieves an average reward of $90$ over $100$ trials. The authors instead seek to prove that the policy will achieve a reward of at least $90$  for any initial condition. By overapproximating the reachable states of the system, they show that the car always receives a total reward greater than $90$ and achieves the goal in less than $115$ steps for a subset of the intial conditions $\hat{p}_0 \in \left[-0.59, -0.4\right]$. 

\subsubsection{Rocket design}
The system under test is a rocket spacecraft with dynamics $m\ddot{p} = f - mge_3$ , where $m > 0$ is the mass, $p(t) \in \mathbf{R}^3$ is the position, and $e_3$ is the unit vector in the z-direction. While it is possible to synthesize optimal trajectories for an idealized model of the system, significant factors such as wind and engine performance (best modeled as random variables) are unaccounted for \cite{blackmore2017autonomous}. Without feedback control, even small uncorrected tracking errors result in loss of the vehicle. In the case of disturbances the authors %
 suggest two approaches: (1) a feedback control law which tracks the optimal trajectory (2) receding horizon model predictive control. The system we consider tracks an optimal trajectory using a feedback control law. Namely, the optimal trajectory is given by the minimum fuel solution to a linearized mode of the dynamics. Specifically, we consider the thrust force discretized in time with a zero-order hold, such that $f_k$ applied for time $t \in [(k-1)h,kh]$ for a time step $h=0.2$. Then, the reference thrust policy solves the following convex optimization problem
 \begin{align*}
 \text{minimize} &\sum_{i=1}^K \| f_k \|_2\\
 \text{such that} ~& p_K = v_K=0, \|f_k\|\le F_{\mathrm{max}},\\
 & v_{k+1}-v_k=\frac{h}{m}f_k-hge_3,\\
 & p_{k+1}-p_k=\frac{h}{2}\left ( v_k + v_{k+1}\right ),\\
 & (p_3)_k \ge 0.5\|((p_1)_k, (p_2)_k)\|_2,
 \end{align*}
where the last constraint is a minimum glide slope and $F_{\mathrm{max} }$ is a maximum thrust value for the nominal thrusters. This results in the thrust profile $f^{\star}$. The booster thrusters correct for disturbances along the flight. The disturbances at every point in time follow a mixture of Gaussians. Namely, we consider 3 wind gust directions, $w_1=(1,1,1)/\sqrt(3)$, $w_2=(0,1,0)$, and $w_3=(1,0,0)$. For every second in time, the wind follows a mixture:
\begin{equation*}
W \sim 	\mc{N}(0,I) + w_1 B + w_2 \hat B  + (1-\hat B)w_3,
\end{equation*}
where $B \sim \text{Bernoulli}(1/3)$ and $\hat B \sim \text{Bernoulli}(1/2)$. This results in 5 random variables for each second, or a total of 100 random variables since we have a 20 second simulation. The wind intensity experienced by the rocket is a linear function of height (implying a simplistic laminar boundary layer): $f_w=CWp_3$ for a constant $C$. Finally, the rocket has a proportional feedback control law for the booster thrusters to the errors in both the position $p_k$ and velocity $v_k$:
\begin{equation*}
f_{\mathrm{feedback,k}} = \text{clip-by-norm}(f^{\star}_k - K_p(p_k-p^{\star}_k) - K_v(v_k-v_k^{\star})).
\end{equation*}
The maximum norm for $\text{clip-by-norm}$ is $aF_{max}$, where $a=1.15$ for Rocket1 and $a=1.1$ for Rocket2, indicating that the boosters are capable of providing $15\%$ or $10\%$ of the thrust of the main engine. 

\subsubsection{Car Racing}

We compare the failure rate of agents solving the car-racing task utilizing the two distinct approaches (\cite{ha2018world} and \cite{tang2020neuroevolution}). The car racing task differs from the other experiments due to the inclusion of a (simple) renderer in the system dynamics. 
At each the step the agent recieves a reward of $ -0.1 +\mathcal{I}_{new tile}(1000/N) - \mathcal{I}_{off track}(100)$ where N is the total number of tiles visited in the track. 
The environment is considered solved if the agent returns an average reward of $900$ over $100$ trials. The search space $P_0$ is the inherent randomness involved with generating a track. The track is generated by selecting 12 checkpoints in polar coordinates, each with radian value uniformly in the interval $[2\pi i/12, 2\pi (i+1)/12)$ for $i = 0, \ldots 11$, and with radius uniformly in the interval $[R/3, R]$, for a given constant value $R$. This results in 24 parameters in the search space. The policies used for testing are described below (with training scripts in the code supplement).
\paragraph{AttentionAgent} \citet{tang2020neuroevolution} utilize a simple self-attention module to select patches from a 96x96 pixel observation. First the input image is normalized then a sliding window approach is used to extract $N$ patches of size $M\times M\times 3$  which are flattened and arranged into a matrix of size $3M^2 \times N$. The self-attention module is used to compute the attention matrix $A$ and importance vector (summation of each column of $A$). A feature extraction operation is applied to the top K elements of the sorted importance vector and the selected features are input to a neural network controller.   
Both the attention module and the controller are trained together via CMA-ES. Together, the two modules contain approximately 4000 learnable parameters. We use the pre-trained model available here: \url{https://github.com/google/brain-tokyo-workshop/tree/master/AttentionAgent}.

\paragraph{WorldModel}
The agent of \citet{ha2018world} first maps a top-down image of the car on track via a variational autoencoder to a latent vector $z$.  Given $z$, the world model $M$ utilizes a recurrent-mixture density network \cite{bishop1994mixture} to model the distribution of future possible states $P(z_{t+1} \mid a_t, z_t, h_t)$. Note that $h_t$, the hidden state of the RNN. Finally, a simple linear controller $C$ maps the concatenation of $z_t$ and $h_t$ to the action, $a_t$. We use the pre-trained model available here: \url{https://github.com/hardmaru/WorldModelsExperiments/tree/master/carracing}.  
\end{document}